\documentclass[10pt,twocolumn,letterpaper]{article}

\usepackage[pagenumbers]{cvpr}      % To produce the REVIEW version
\usepackage{graphicx}
\usepackage{booktabs} 
\usepackage{multirow}
\usepackage{array} % Import necessary packages
\usepackage{subcaption}
\usepackage{xcolor,colortbl}
\usepackage{multirow}
\usepackage{pifont}
\usepackage{amssymb}
\usepackage{amssymb}
\usepackage{bbding}
\usepackage{float}
\usepackage{siunitx}
\usepackage{makecell}

\definecolor{deepgreen}{RGB}{0,200,0} % Define a deeper green color
\definecolor{cvprblue}{rgb}{0.21,0.49,0.74}
\usepackage[pagebackref,breaklinks,colorlinks,allcolors=cvprblue]{hyperref}

\newcommand{\Ariaemoji}{\includegraphics[height=1.3\fontcharht\font`\B]{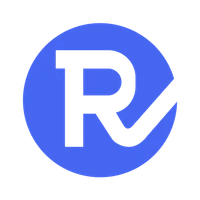}}
\newcommand{\Qwenemoji}{\includegraphics[height=1.3\fontcharht\font`\B]{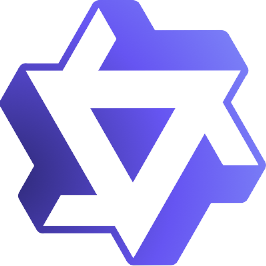}}
\newcommand{\Googleemoji}{\includegraphics[height=1.3\fontcharht\font`\B]{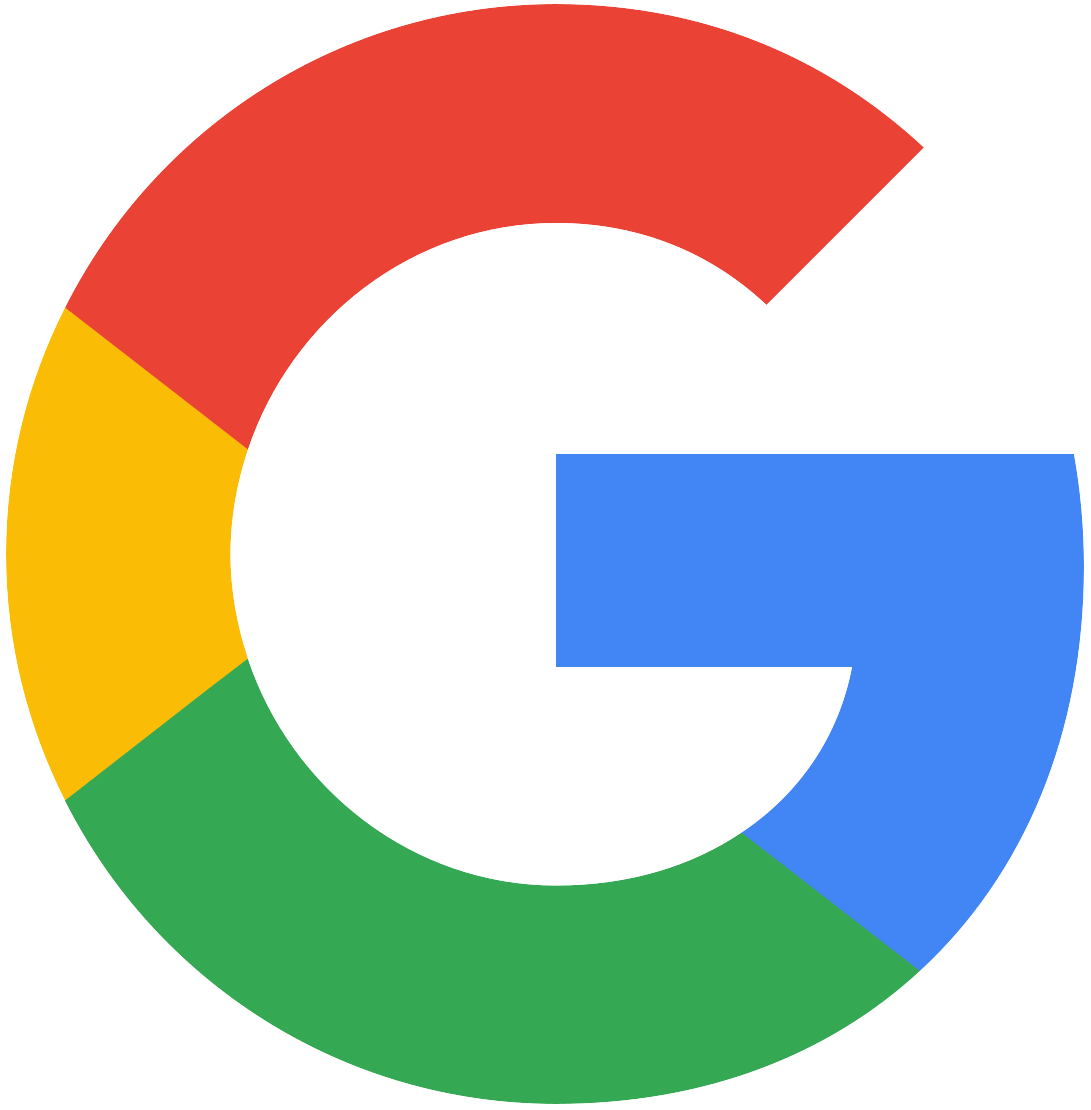}}
\newcommand{\Openaiemoji}{\includegraphics[height=1.2\fontcharht\font`\B]{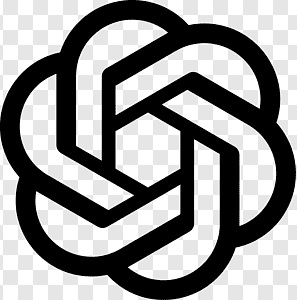}}
\newcommand{\llavanextmoji}{\includegraphics[height=1.5\fontcharht\font`\B]{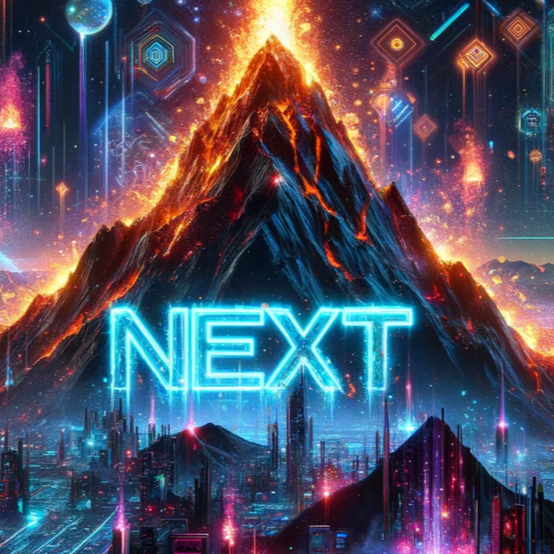}}

\title{VideoAutoArena: An Automated Arena for Evaluating Large Multimodal Models in Video Analysis through User Simulation}

\author{%
  Ziyang Luo$^{1,2}$,~
  Haoning Wu$^{4}$,~
  Dongxu Li$^{5}$,~
  Jing Ma$^{2}$,~
  Mohan Kankanhalli$^{3}$,~
  Junnan Li$^{1}$\\[4pt]
  $^1$Salesforce AI Research, $^2$Hong Kong Baptist University, $^3$National University of Singapore\\
  $^4$Nanyang Technological University, $^5$The Australian National University
}

\begin{document}

\twocolumn[{
\renewcommand\twocolumn[1][]{#1}
\maketitle
\begin{center}
    \vspace{-1.0cm}
    \includegraphics[width=\linewidth]{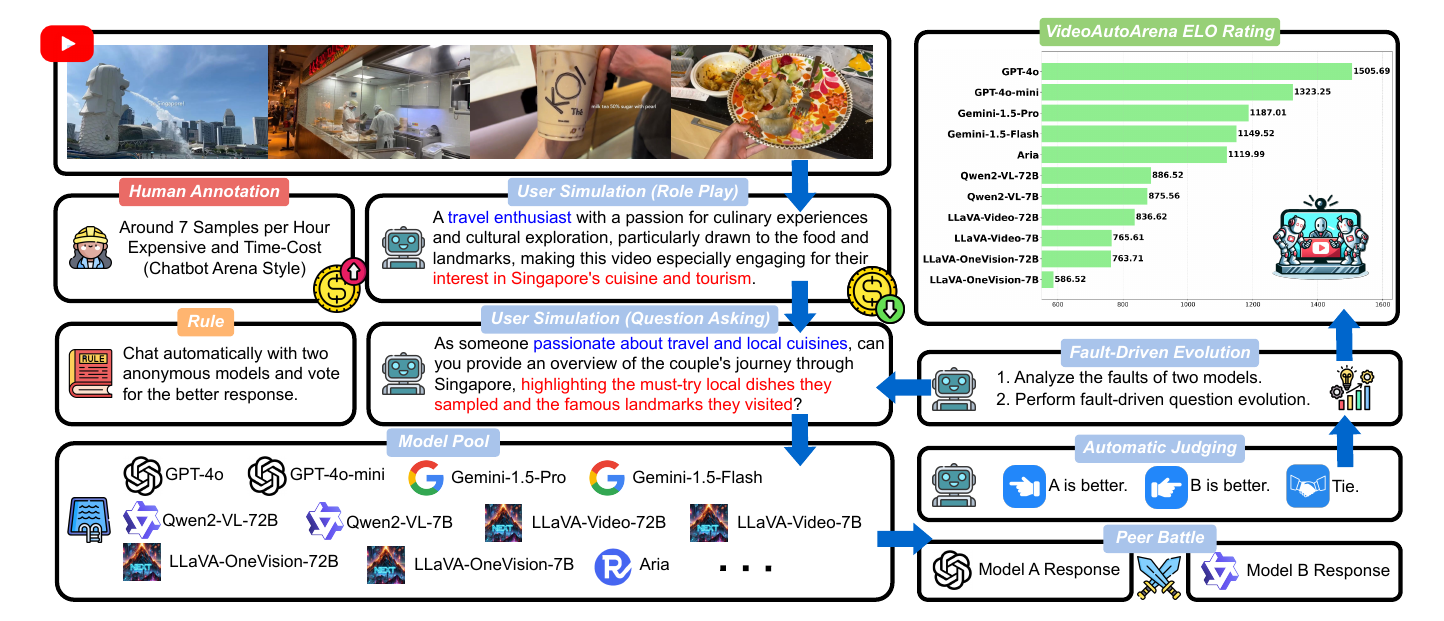}
    \captionsetup{type=figure}
    \vspace{-0.2cm}
    \caption{An overview of our \textbf{VideoAutoArena}, where we leverage LMMs for user simulation to automatically evaluate LMMs in video analysis, offering an efficient alternative to costly and time-consuming human annotations, distinct from platforms like LMSYS Chatbot Arena~\cite{ChatbotArena} and WildVision Arena~\cite{WildVision}. In this figure, we showcase 4 sampled frames from a Singapore travel vlog video.}
    % \vspace{-0.1cm}
    \label{fig:intro}
\end{center}
}]

\let\thefootnote\relax\footnotetext{$^\diamondsuit$Project Page: \href{https://videoautoarena.github.io/}{\texttt{https://videoautoarena.github.io/}}}

\begin{abstract}
Large multimodal models (LMMs) with advanced video analysis capabilities have recently garnered significant attention. However, most evaluations rely on traditional methods like multiple-choice question answering in benchmarks such as VideoMME and LongVideoBench, which are prone to lack the depth needed to capture the complex demands of real-world users.
To address this limitation—and due to the prohibitive cost and slow pace of human annotation for video tasks—we introduce \textbf{VideoAutoArena}, an arena-style benchmark inspired by LMSYS Chatbot Arena's framework, designed to automatically assess LMMs' video analysis abilities. VideoAutoArena utilizes user simulation to generate open-ended, adaptive questions that rigorously assess model performance in video understanding. The benchmark features an automated, scalable evaluation framework, incorporating a modified ELO Rating System for fair and continuous comparisons across multiple LMMs.
To validate our automated judging system, we construct a ``gold standard" using a carefully curated subset of human annotations, demonstrating that our arena strongly aligns with human judgment while maintaining scalability. Additionally, we introduce a fault-driven evolution strategy, progressively increasing question complexity to push models toward handling more challenging video analysis scenarios.
Experimental results demonstrate that VideoAutoArena effectively differentiates among state-of-the-art LMMs, providing insights into model strengths and areas for improvement. To further streamline our evaluation, we introduce \textbf{VideoAutoBench} as an auxiliary benchmark, where human annotators label winners in a subset of VideoAutoArena battles. We use GPT-4o as a judge to compare responses against these human-validated answers. Together, VideoAutoArena and VideoAutoBench offer a cost-effective, and scalable framework for evaluating LMMs in user-centric video analysis.
\end{abstract}
\section{Introduction}\label{sec:intro}

\begin{table*}[t]
\small
\centering
\begin{tabular}{lcccccc}
\toprule
\textbf{Benchmark} & \textbf{Venue} & \textbf{Long Video Included} & \textbf{User-Centric} & \textbf{Scalable} & \textbf{Open-Ended} & \textbf{Automated} \\
\midrule
\textbf{MVBench}~\cite{MVBench} & CVPR 24 & \textcolor{red}{\ding{55}} & \textcolor{red}{\ding{55}} & \textcolor{red}{\ding{55}} & \textcolor{red}{\ding{55}} & \textcolor{deepgreen}{\checkmark} \\
\textbf{MLVU}~\cite{MLVU} & Arxiv 24 & \textcolor{deepgreen}{\checkmark} & \textcolor{red}{\ding{55}} & \textcolor{red}{\ding{55}} & \textcolor{red}{\ding{55}} & \textcolor{deepgreen}{\checkmark} \\
\textbf{LVBench}~\cite{LVBench} & Arxiv 24 & \textcolor{deepgreen}{\checkmark} & \textcolor{red}{\ding{55}} & \textcolor{red}{\ding{55}} & \textcolor{red}{\ding{55}} & \textcolor{deepgreen}{\checkmark} \\
\textbf{VideoMME}~\cite{Video-MME} & Arxiv 24 & \textcolor{deepgreen}{\checkmark} & \textcolor{red}{\ding{55}} & \textcolor{red}{\ding{55}} & \textcolor{red}{\ding{55}} & \textcolor{deepgreen}{\checkmark} \\
\textbf{LongVideoBench}~\cite{LongVideoBench} & NeurIPS 24 & \textcolor{deepgreen}{\checkmark} & \textcolor{red}{\ding{55}} & \textcolor{red}{\ding{55}} & \textcolor{red}{\ding{55}} & \textcolor{deepgreen}{\checkmark} \\
\textbf{WildVision Video Arena}~\cite{WildVision} & NeurIPS 24 & \textcolor{violet}{\textbf{?}} & \textcolor{deepgreen}{\checkmark} & \textcolor{red}{\ding{55}} & \textcolor{deepgreen}{\checkmark} & \textcolor{red}{\ding{55}} \\
\textbf{VideoAutoArena (Ours)} & CVPR 25 & \textcolor{deepgreen}{\checkmark} & \textcolor{deepgreen}{\checkmark} & \textcolor{deepgreen}{\checkmark} & \textcolor{deepgreen}{\checkmark} & \textcolor{deepgreen}{\checkmark} \\
\bottomrule
\end{tabular}
\caption{Comparison of recent popular benchmarks for video analysis. WildVision video data are not yet publicly available.}
\vspace{-0.4cm}
\label{tab:evaluation-comparison}
\end{table*}

Recently, large multimodal models (LMMs) with advanced video understanding capabilities, such as GPT-4o~\cite{GPT-4o}, Gemini-1.5-Pro~\cite{Gemini,Gemini1.5}, Aria~\cite{Aria}, Qwen2-VL~\cite{Qwen2-VL}, and LLaVa-Video~\cite{LLaVA-OneVision,Llava-Video}, have garnered significant attention within the multimodal community. These models represent a major shift in the capabilities of artificial intelligence by extending beyond traditional image-based LMMs~\cite{BLIP-2,InstructBLIP,llava,GPT-4V} to encompass complex video inputs. Unlike their predecessors, which primarily focused on static visual data, these models are designed to handle dynamic, time-variant data, making them well-suited for processing lengthy and intricate video sequences. By leveraging language-based instructions from users, these LMMs can efficiently conduct a range of video analysis tasks, from understanding fine-grained scene details~\cite{Aria} to comprehending overarching narrative structures~\cite{GPT-4o}.

To better evaluate the video analysis capabilities of these models, recent works have introduced several widely used benchmarks, including MVBench~\cite{MVBench}, VideoMME~\cite{Video-MME}, and LongVideoBench~\cite{LongVideoBench}. These benchmarks typically share a common feature: they pre-define core video analysis skills, such as object recognition in a single frame and action reasoning across a sequence of frames. They adopt an ability-centric approach, often using multiple-choice questions to assess performance. While these benchmarks have significantly contributed to the development of LMMs, they place limited emphasis on the types of questions real users might ask when seeking assistance with video analysis. In contrast, practical video analysis scenarios are far more complex and diverse in their requirements~\cite{ChatbotArena}.

To address this gap, we can draw inspiration from the evaluation methods used for large language models (LLMs)~\cite{GPT3,LLaMA,LLaMA2,llama3,DBLP:journals/corr/abs-2405-20267}. Traditional language benchmarks, such as MMLU~\cite{MMLU}, IFEval~\cite{IFEval}, HumanEval~\cite{HumanEval}, and GSM8k~\cite{GSM8k}, also suffer from limited alignment with real user interactions~\cite{zhang-etal-2024-naturalcodebench,ChatbotArena,wildchat}. To mitigate this issue, platforms like LMSYS Chatbot Arena~\cite{ChatbotArena} have been introduced, providing an open, crowdsourced platform for evaluating LLMs based on human preferences. Chatbot Arena employs a pairwise comparison approach, collecting feedback from a diverse set of users, ensuring that evaluation questions are generated by real users. This approach overcomes many of the challenges posed by previous benchmarks and has become one of the most widely adopted methods for evaluating LLMs.

One straightforward solution is to adapt this idea directly to LMMs for video analysis. Indeed, recent work, such as WildVision Arena~\cite{WildVision}, has attempted to do so. However, an analysis of the WildVision Video Arena leaderboard reveals certain challenges with this approach. The video arena has received only 256 votes across 11 models, with each model participating in an average of just 23 battles since its release approximately 6 months ago.\footnote{The numbers are recorded on Nov. 14, 2024 at \url{https://huggingface.co/spaces/WildVision/vision-arena}.} This limited number of battles can likely be attributed to the increased complexity involved in formulating questions for video-based tasks. Unlike language and image data, where questions can be quickly generated or verified within seconds, videos are typically longer and contain richer, more complex contexts. Annotators must spend significantly more time watching and understanding the video content before formulating high-quality questions. In our preliminary attempts, we hired annotators to complete this task, and the time cost proved significant, with \textbf{a maximum output of only 7 samples annotated per hour}. This time investment hinders the scalability of generating high-quality questions for LMMs, a crucial factor for ensuring the effectiveness of arena-style evaluations.

% VideoAutoArena and results
To address the limitations of existing video analysis benchmarks, we propose \textbf{VideoAutoArena}, a fully automated, arena-style evaluation method for LMMs. Unlike human-driven platforms, VideoAutoArena leverages LMM agents for user simulation and preference selection, eliminating the need for costly human annotators and enabling scalable, efficient evaluations. The framework also integrates fault-driven hard prompt evolution, which generates progressively challenging questions based on model performance, ensuring more rigorous testing. By simulating real-world user behavior, VideoAutoArena bridges the gap between ability-centric evaluations and practical application demands. Our human preference experiments show that \textbf{84.20\%} of the time, questions in VideoAutoArena better mimic real-world user question styles compared to VideoMME and LongVideoBench. Additionally, \textbf{87.29\%} of the time, our automatic judging aligns with human preference selection.

Experiments on 11 well-known proprietary and open-source LMMs reveal that open-source models still lag behind the SOTA closed-source model GPT-4o in video analysis, with a significant performance gap (-385.7). This gap is notably larger than those observed in traditional multiple-choice question-answering benchmarks. The disparity becomes even more pronounced as video length increases or the difficulty of the questions rises. Furthermore, when focusing on user-background relevance and helpfulness, the performance gap widens further. These findings highlight how our benchmark offers a user-centric perspective, providing valuable insights for the development of LMMs.

% VideoAutoBench and results
To complement VideoAutoArena, we also introduce \textbf{VideoAutoBench}, a streamlined benchmark designed for faster, more accessible evaluation of LMMs in video analysis. VideoAutoBench leverages a curated subset of battles from VideoAutoArena, where human annotators have selected the winning model responses. Using GPT-4o for automatic judging, VideoAutoBench compares model answers against these human-selected and rejected responses, providing an efficient, cost-effective assessment method. Our results show that the rankings from VideoAutoBench align closely with those from VideoAutoArena, with a significant gap between SOTA closed-source and open-source LMMs, underscoring the benchmark's challenge.
\section{Related Work}

% Video LMMs
LMMs with advanced video understanding capabilities have garnered significant research attention. For the closed-source models, GPT-4o~\cite{GPT-4o} and Google's Gemini-1.5~\cite{Gemini,Gemini1.5} demonstrate SOTA video analysis performance. Meanwhile, the open-source community has made notable strides~\cite{VideoChat,Video-LLaMA,ShareGPT4Video,Video-LLaVA,VILA,PLLaVA,VideoLLaMA2,LongVILA,LongVA,Oryx,minicpm,VITA,Kangaroo,Video-CCAM,InternVL,Chat-UniVi,ST-LLM,BLIP-3-Video,VTimeLLM,VideoLLM,Otter,mPLUG-Owl3,Macaw-LLM,Valley,Merlin}. Notably, the LLaVa series~\cite{llava} has been updated to the LLaVa-Video~\cite{Llava-Video} and LLaVa-OneVision models~\cite{LLaVA-OneVision}, along with the release of all training data. The Qwen-VL model~\cite{Qwen-VL} has also been upgraded to the Qwen2-VL version~\cite{Qwen2-VL}, and the first open-source multimodal mixture-of-experts (MoE) model, Aria~\cite{Aria}, has recently been introduced.
These contributions have significantly narrowed the gap between closed-source and open-source models in video understanding. To accelerate the development of LMMs in video analysis, the establishment of more comprehensive benchmarks is essential.

% Video Benchmark 
In the early phase, researchers primarily relied on benchmarks featuring short videos~\cite{DBLP:journals/corr/abs-2311-14906,AGQA,DBLP:conf/iccv/GoyalKMMWKHFYMH17,DBLP:conf/nips/PatrauceanS0RMB23}, such as MSVD-QA~\cite{MSVD}, MSRVTT-QA~\cite{MSVD}, NExT-QA~\cite{NExT-QA}, and MVBench~\cite{MVBench}. However, these benchmarks have limitations due to their short video durations, averaging less than 50 seconds. This brevity restricts their ability to comprehensively evaluate the temporal understanding capabilities of LMMs, thereby hindering further advancements in LMMs development. To address these limitations, benchmarks like ActivityNet-QA~\cite{ActivityNet-QA} and EgoSchema~\cite{EgoSchema} have extended video durations to approximately 180 seconds on average. More recently, research has introduced even more comprehensive benchmarks~\cite{MoVQA,TempCompass,LVBench}. For instance, MovieChat-1K~\cite{MovieChat} assesses LMMs using movie videos with an average duration of 500 seconds, while LongVideoBench~\cite{LongVideoBench} focuses on long-context interleaved evaluation with an average duration of 473 seconds. Additionally, MLVU~\cite{MLVU} presents a LMMs' ability-centric benchmark, featuring substantial extensions of video lengths. Furthermore, VideoMME~\cite{Video-MME} introduces a highly comprehensive benchmark that includes short, medium, and long videos, further enhancing the evaluation of LMMs' temporal understanding abilities. As discussed in the introduction, most current benchmarks are limited by multiple-choice questions that diverge from real user interaction. To address this, we introduce \textbf{VideoAutoArena}, which evaluates LMMs through open-ended, simulated human questions. Table~\ref{tab:evaluation-comparison} highlights the differences between our benchmark and other recent video-based benchmarks.
\section{VideoAutoArena}\label{sec:arena}
\subsection{Overview}

\begin{figure}
    \centering
    \includegraphics[width=\linewidth]{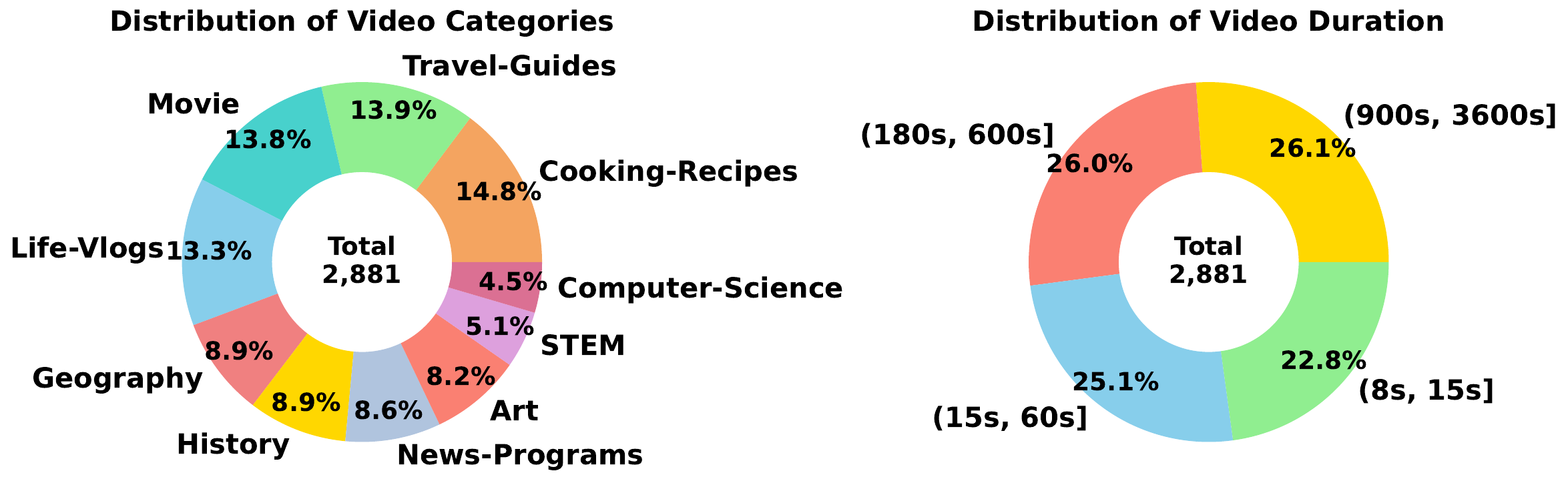}
    \vspace{-0.5cm}
    \caption{Video statistics by category and duration.}
    \vspace{-0.4cm}
    \label{fig:video_stat}
\end{figure}

As illustrated in Figure~\ref{fig:intro}, the \textbf{VideoAutoArena} pipeline consists of four core components: user simulation, peer battles, automatic judging, and fault-driven evolution. Initially, an agent reviews a video to identify user personas likely to be interested in the content. Adopting one of these personas, the agent formulates a relevant question about the video. Two randomly selected models then engage in peer battles to respond to the question. A judging agent determines which model provides the better response, while an analysis agent evaluates the responses, performs fault analysis, and generates progressively challenging questions to further assess the models' capabilities. To demonstrate the effectiveness of our VideoAutoArena, we use 2,881 videos from LongVideoBench, with an average duration of 479 seconds. These videos are categorized into four duration ranges—(8s, 15s], (15s, 60s], (180s, 600s], (900s, 3600s]—and 10 categories: Movie, Life Vlogs, Geography, History, News Programs, Art, STEM, Computer Science, Cooking Recipes, and Travel Guides. The data distribution are provided in Figure~\ref{fig:video_stat}. Our benchmark is not restricted to specific videos, and new videos can be easily incorporated into the evaluation pipeline.

\subsection{User Simulation}

Ideally, real user queries would offer the direct evaluation of LMM performance in real life video analysis. However, given the complex contextual demands of video content, \textbf{human annotation is expensive and time-cost}. To address this, \textbf{VideoAutoArena} adopts user simulation with role-play, with SOTA LMMs acting as agents to generate realistic, user-centeric questions and preferences, enabling a more practical evaluation. In language-based roly-play~\cite{Personahub,Kong2024FromGT,DBLP:journals/corr/abs-2405-02957,DBLP:journals/corr/abs-2408-08089}, there is typically no need to consider external context, such as videos, allowing role-play to freely generate diverse questions. In video analysis, however, questions are constrained by the video content. To address this, we introduce a novel role-play method called \textbf{video content-constrained persona generation}. Given a video, agents first identify the types of users likely to be interested in it, defining three user types: (1) users with backgrounds highly relevant to the video; (2) users with moderately relevant backgrounds; and (3) users with unrelated backgrounds who encounter the video by chance. This relevance-based categorization aims to emulate real users with varied backgrounds seeking assistance from LMMs for video analysis. Once user types are established, agents adopt these personas to generate persona-specific questions for detailed video analysis. This user-centric process sets our evaluation method apart from previous ability-centric benchmarks. The prompts used are provided in the Appendix~\ref{app:user}.

\subsection{Peer Battles and Fault-Driven Evolution}

Once the role-play agent generates a question for a specific video, we initiate peer battles between two anonymous models, following the Chatbot Arena style. Two models are randomly chosen from the model pool, presented with the same video and question, and asked to generate responses. Our goal is to fairly compare the quality of these responses to determine which LMM provides a better answer.

Similar to the concept of Hard Prompts in Chatbot Arena~\cite{ArenaAutoHard}, incorporating more challenging questions can further push the boundaries of evaluating the abilities of current LMMs in video analysis. Thus, we aim to create a harder question set for evaluation purposes. Unlike Chatbot Arena, which can directly source hard prompts from millions of real user queries, our approach is limited to the prompts generated by a user simulator. To derive harder questions, we draw inspiration from the famous instruction synthesis method, Evol-Instruct~\cite{WizardLM,WizardCoder}, which evolves existing questions into more complex ones using predefined heuristic strategies. However, because Evol-Instruct generates questions based on a similar prompt structure for each evolution, it encounters limitations in question diversity. To address this, we introduce a \textbf{fault-driven question evolution} strategy that iteratively increases question complexity. Rather than relying on isolated prompts, each new evolution generates questions based on the results of the previous model battle. This approach creates a more adaptive and progressively challenging environment for the models, pushing them to respond to increasingly complex questions.

In this framework, a response analysis agent initially reviews responses from two competing models, identifying specific faults and performance weaknesses. Based on this assessment, role-play agents then generate tailored questions aimed at probing these weaknesses, making the question synthesis process progressively more fault-driven. After a new question is generated, a complexity evaluation agent assesses its difficulty. If the new question receives a higher overall complexity score than the previous one, it is retained for subsequent model battles. This iterative approach establishes a rigorous testing environment, challenging models with increasingly complex and contextually nuanced tasks, thus providing a deeper evaluation of each model's video understanding capabilities. The prompts used are provided in the Appendix~\ref{app:evol}.

\begin{figure*}
    \centering
    \includegraphics[width=\linewidth]{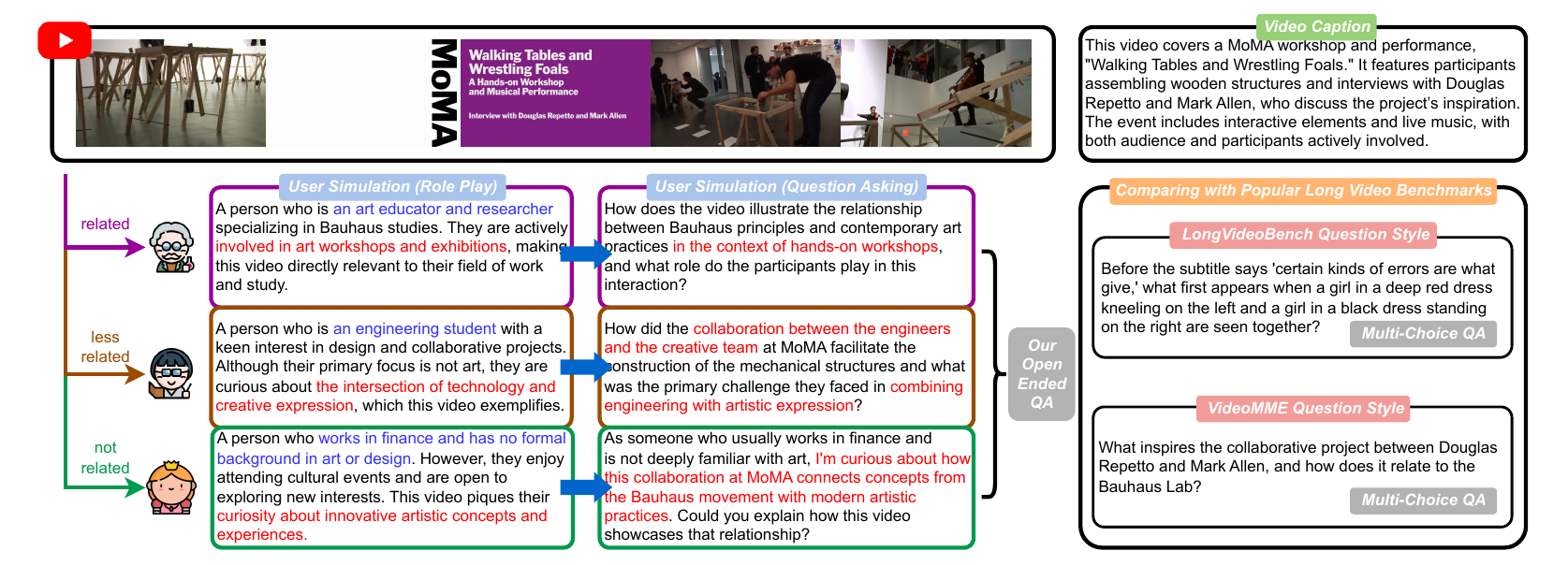}
    \vspace{-0.5cm}
    \caption{Examples of synthesized personas with three levels of relevance and corresponding synthesized questions. We also compare the style of our questions with those in popular long-video benchmarks, including LongVideoBench and VideoMME.}
    \vspace{-0.4cm}
    \label{fig:examples}
\end{figure*}
\subsection{Judging and Ranking}

A key aspect of arena-style evaluation is determining the winner in each model comparison. In Chatbot Arena, human annotators directly express their preferences, but in \textbf{VideoAutoArena}, human annotation is costly and difficult to scale due to the time-intensive nature of video analysis. To address this, we aim to automate the judgment process. Drawing inspiration from automated judging benchmarks like Arena-Auto-Hard~\cite{ArenaAutoHard} and MT-Bench~\cite{MTBench}, we first define our judging standards as follows:
\begin{enumerate}
\item \textbf{Instruction Following}: The response should closely adhere to the user’s instructions, ensuring it directly addresses the specified task.
\item \textbf{Accuracy}: The response must utilize information from the video accurately, avoiding fabrication or misquotation. It should maintain factual correctness, avoid hallucinations, and demonstrate contextual coherence with precise terminology and knowledge.
\item \textbf{Relevance}: The response should consider the user’s background information and needs, providing a comprehensive, detailed answer that addresses the question directly without straying off-topic. Responses should be thorough, offering multiple perspectives where relevant.
\item \textbf{Helpfulness}: The response should provide valuable information to aid the user in understanding or solving their issue, avoiding irrelevant or vague content.
\end{enumerate}
Based on these standards, we adopt the LMM-as-a-Judge paradigm~\cite{ArenaAutoHard,MTBench,WildVision} to automatically determine the better response between two models. The prompts used are provided in the Appendix~\ref{app:judge}.

After obtaining the automatic judging results, we utilize the ELO Rating System~\cite{ELO} to establish a dynamic evaluation platform that ranks LMMs through statistical modeling based on direct pairwise comparisons. Here, we briefly explain the Online ELO Rating and statistical estimation methods. The Online ELO Rating system calculates the probability that model \(i\) will win against model \(j\) using their respective ratings, \(R_i\) and \(R_j\), where \(i, j \in N\). For each comparison, we define a binary outcome \(Y_{ij}\), where \(Y_{ij} = 1\) if model \(i\) wins and \(Y_{ij} = 0\) otherwise. The probability is computed as follows:

\[
P(Y_{ij}=1) = \frac{1}{1 + 10^{(R_j - R_i)/\alpha}},
\]

where \(\alpha = 400\) is the scaling factor in the ELO computation. After each comparison, player ratings are updated by:

\[
R_i' = R_i + K \times (S(i, j) - P(Y_{ij} = 1)),
\]

where \(S(i, j)\) represents the actual outcome (1 for a win, 0.5 for a tie, and 0 for a loss). Higher-rated players gain fewer points if they win and lose more if defeated, while lower-rated players experience the reverse. Since ELO updates are sensitive to comparison order, we employ the Bradley–Terry model~\cite{Bradley1952RankAO} for stable statistical estimation, as in Chatbot Arena.

The Bradley–Terry model refines ELO ratings through a logistic regression model using maximum likelihood estimation. For \(N\) models with pairwise comparisons, where \(W_{ij}\) is the count of times model \(i\) wins over \(j\), the log-likelihood function for all comparisons is:

\[
\mathcal{R} = \sum_{i, j \in N, i \neq j} W_{ij} Y_{ij} P(Y_{ij} = 1),
\]

where \(\mathcal{R} = \{R_1, \dots, R_N\}\) represents each model’s rating. Since this approach doesn’t accommodate ties directly, we split all tie votes, counting half as wins for model \(i\) (\(Y_{ij} = 1\)) and the other half as wins for model \(j\) (\(Y_{ij} = 0\)). This ensures balanced ranking and fair statistical estimation across all model comparisons.

\subsection{Experiments}

\textbf{Setup.} To demonstrate the effectiveness of VideoAutoArena, we evaluate 11 SOTA LMMs, including GPT-4o/mini, Gemini-1.5-Pro/Flash, Aria, Qwen2-VL-72B/7B, LLaVa-Video-72B/7B, and LLaVa-OneVision-72B/7B, all of which have shown strong performance on the VideoMME.\footnote{Claude-Series LMMs were excluded from our experiments due to limited support for long video input (maximum 20 frames).} For response generation, each video was uniformly sampled to provide 64 frames as input. Since most of these LMMs do not support audio, the audio track was converted to subtitles and combined with the question as input. For automatic judging, each video was sampled to provide 128 frames and combined with subtitles. Additional experimental details are provided in the Appendix~\ref{app:exp}.

\begin{figure}
    \centering
    \begin{subfigure}[b]{\linewidth}
    \centering
        \includegraphics[width=0.7\textwidth]{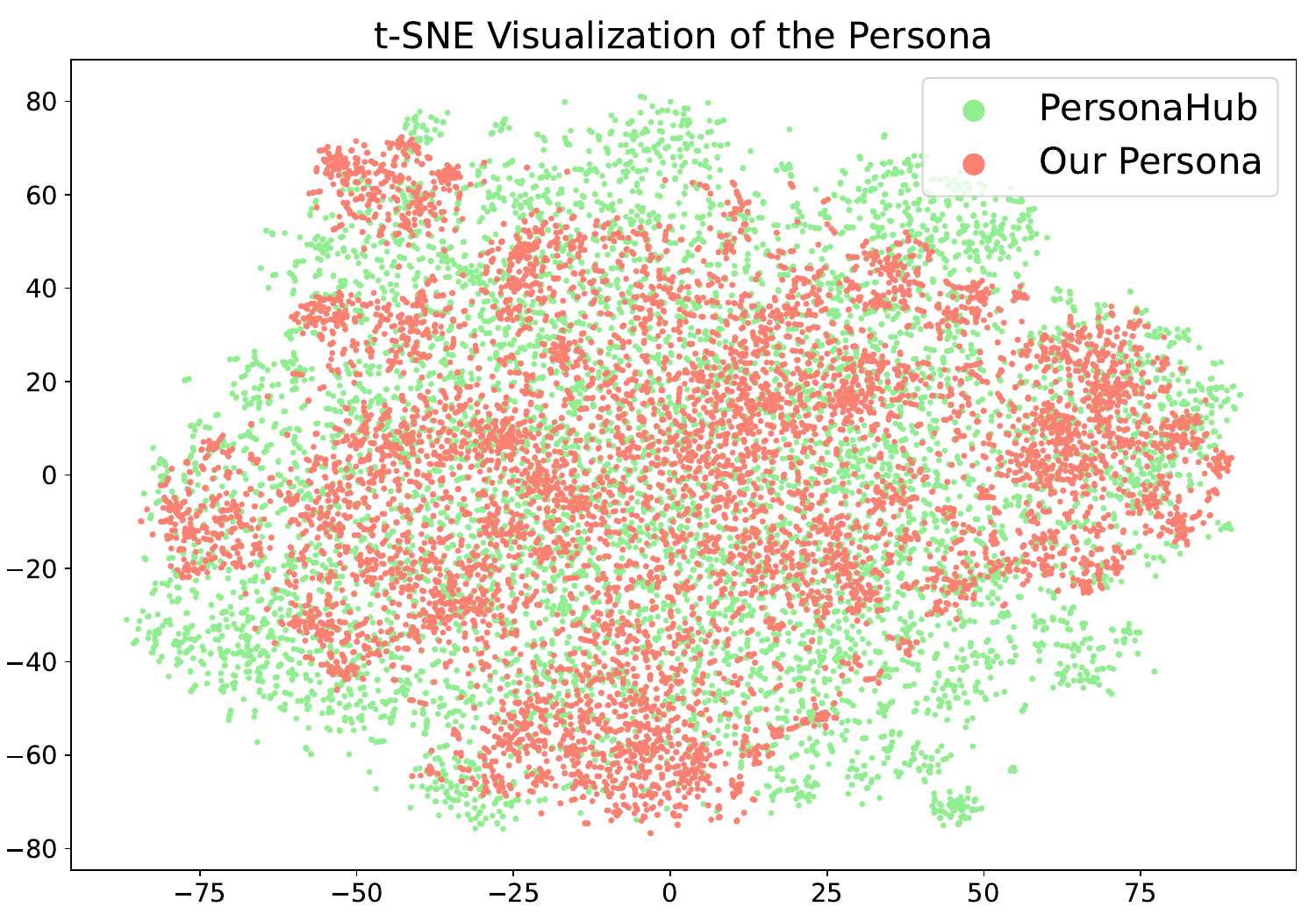}
        \caption{Visualization of persona distribution.}
        \label{fig:visual_persona}
    \end{subfigure}
    \hfill
    \begin{subfigure}[b]{\linewidth}
    \centering
        \includegraphics[width=0.7\textwidth]{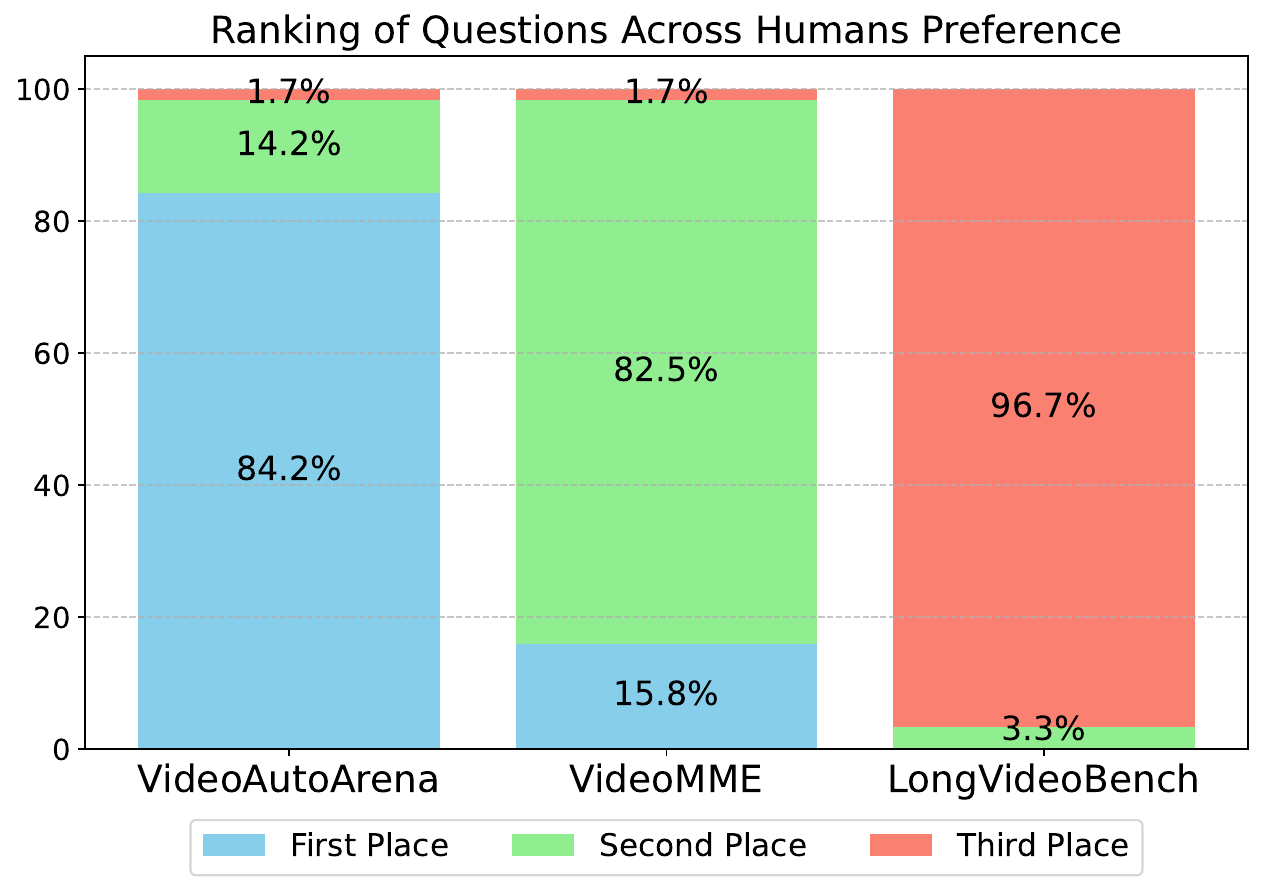}
        \caption{Humans preference ranking.}
        \label{fig:rank_question}
    \end{subfigure}
    \vspace{-0.5cm}
    \caption{Our user simulation offers diverse personas and more effectively mirrors real-world users' question styles.}
    \vspace{-0.4cm}
    \label{fig:two_plots}
\end{figure}
\textbf{User Simulation and Diversity.} In our experiments, we generated an average of three personas per video across 2.9k videos, resulting in about 8.6k unique personas. As shown in Figure~\ref{fig:examples}, we showcases examples of three personas with varying levels of relevance to a given video. Each persona includes motivations or reasons for their interest in the video, enabling more persona-specific question generation. Moreover, in Figure~\ref{fig:visual_persona}, we compare the distribution of our synthesized personas with those from PersonaHub~\cite{Personahub}, which includes diverse personas automatically curated from web. To map the distribution, we used one of the SOTA sentence embedding models, \texttt{gte-large-en-v1.5}~\cite{gte}, to encode each persona description into vectors, then applied t-SNE~\cite{tsne} for dimensionality reduction. The results show that our synthesized personas achieve diversity comparable to that in PersonaHub. Notably, our personas are constrained by video content, while PersonaHub personas are synthesized without such constraints, highlighting that our personas effectively simulate a range of realistic user backgrounds. Additional examples of synthesized personas are provided in Appendix~\ref{app:example_persona}.

After generating personas, our role-play agent adopts each persona to ask relevant questions about the video content, seeking help from AI assistants. In our experiments, we synthesized one question per unique personas for 2.9k videos. To assess how well our questions mimic real-world user queries, we randomly selected 120 questions and compared them with question styles from LongVideoBench and VideoMME. Since our videos are from LongVideoBench, we used the same set for consistency. However, since VideoMME uses different videos, we applied a style transfer method, adapting VideoMME questions to synthesize similar-style questions based on our benchmark's videos. This allowed a fair comparison across the three benchmarks on identical video samples. To evaluate the naturalness of each benchmark's question style, we conducted a blind ranking task, asking three skilled annotators to rank which questions best mimic real-world user queries, with Rank 1 as the best. Annotation guidelines are detailed in the Appendix~\ref{app:annotation_rank}. As shown in Figure~\ref{fig:rank_question}, for 84.2\% of the comparisons, VideoAutoArena's questions ranked first in mimicking real-world user question styles. This result highlights how our benchmark brings a unique perspective to evaluating LMMs for video analysis. Figure~\ref{fig:examples} further provides example questions to illustrate the style of our questions in mimicking real-world users. In Appendix~\ref{app:example_question}, we also include more synthesized question examples.

\begin{table*}
    \small
    \centering
    \begin{tabular}{lccccccc}
        \toprule
        \textbf{Models} & \textbf{Size} & \textbf{ELO} & \textbf{Win Rates}& (\SI{8}{\second}, \SI{15}{\second}] & (\SI{15}{\second}, \SI{60}{\second}] & (\SI{180}{\second}, \SI{600}{\second}] & (\SI{900}{\second}, \SI{3600}{\second}]\\
        \midrule
        \rowcolor{pink!50}
        \multicolumn{8}{c}{\textit{Proprietary Models}}\\
        \Openaiemoji{}~\textbf{GPT-4o} & - & \textbf{1505.69} & \textbf{89.19} & \textbf{1447.86} & \textbf{1449.59} & \textbf{1575.34} & \textbf{1552.23}\\
        \Openaiemoji{}~\textbf{GPT-4o-mini} & - & 1323.25 & 76.90 & 1293.27 & 1343.28 & 1327.75 & 1349.29\\
        \Googleemoji{}~\textbf{Gemini-1.5-Pro}& - & 1187.01 & 65.11 & 1247.65 & 1171.82 & 1263.58 & 1291.64\\
        \Googleemoji{}~\textbf{Gemini-1.5-Flash}& - & 1149.52 & 62.07 & 1081.58 & 1131.27 & 1140.07 & 1260.36\\
        \midrule
        \rowcolor{green!30}
        \multicolumn{8}{c}{\textit{Open-Source Models}}\\
        \Ariaemoji{}~\textbf{Aria} & 8$\times$3.5B & \textbf{1119.99} & \textbf{59.54} & \textbf{1147.45} & \textbf{1273.77} & \textbf{1110.67} & \textbf{1111.40}\\
        \Qwenemoji{}~\textbf{Qwen2-VL} & 72B & 886.52 & 35.61 & 985.46 & 928.23 & 829.65 & 826.56\\
        \Qwenemoji{}~\textbf{Qwen2-VL} & 7B & 875.56 & 34.90 & 969.28 & 859.33 & 850.30 & 829.21\\
        \llavanextmoji{}~\textbf{LLaVA-Video} & 72B & 836.62 & 30.25 & 796.90 & 850.12 & 827.88 & 782.55\\
        \llavanextmoji{}~\textbf{LLaVA-Video} & 7B & 765.61 & 23.52 & 672.35 & 736.14 & 759.15 & 721.78\\
        \llavanextmoji{}~\textbf{LLaVA-OneVision} & 72B & 763.71 & 23.11 & 731.50 & 710.64 & 759.29 & 741.80\\
        \llavanextmoji{}~\textbf{LLaVA-OneVision} & 7B & 586.52 & 9.86 & 626.70 & 545.82 & 556.31 & 533.18\\
        \bottomrule
    \end{tabular}
    \caption{Our \textbf{VideoAutoArena} Leaderboard. We show the overall ELO ratings and win rates within four different video lengths.}
    \vspace{-0.4cm}
    \label{tab:ELO}
\end{table*}
\begin{figure}
    \centering
    \includegraphics[width=0.7\linewidth]{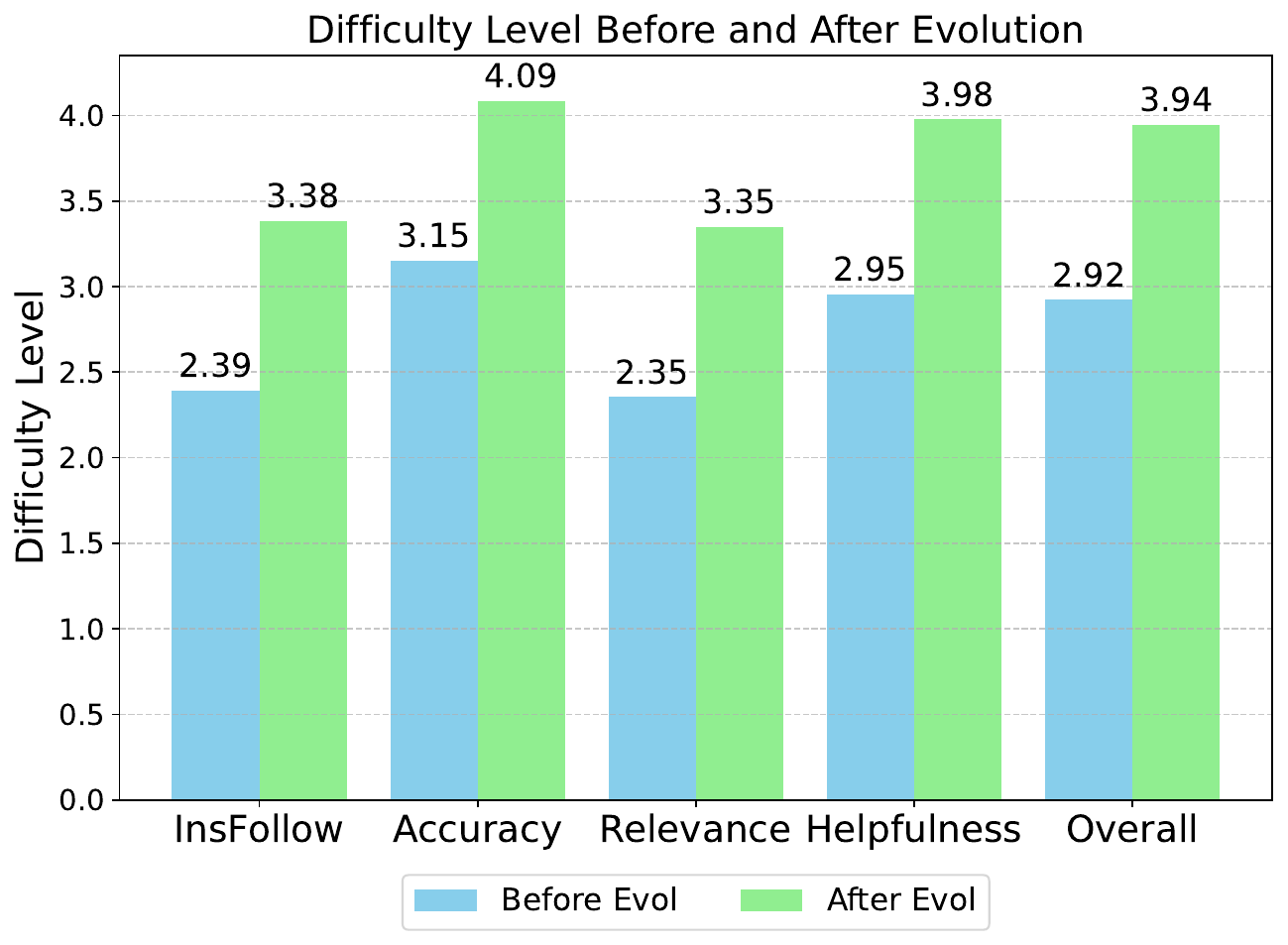}
    \vspace{-0.3cm}
    \caption{Our fault-driven evolution strategy generates increasingly challenging questions for video analysis.}
    \vspace{-0.4cm}
    \label{fig:question_diff}
\end{figure}
\textbf{Question Evolution.} To analyze the hard prompts generated by our fault-driven evolution, we perform an automated complexity assessment of the synthesized questions. Using the SOTA LMM, GPT-4o, we evaluate the difficulty level of these questions on a 1–5 scale before and after evolution. We further break down the difficulty into 5 categories—instruction following, accuracy, relevance, helpfulness and overall—based on our automatic judging standards. As shown in Figure~\ref{fig:question_diff}, the evolved questions consistently achieve higher difficulty scores across all categories, demonstrating the effectiveness of our evolution strategy. The analysis prompts are included in Appendix~\ref{app:comp}.

\begin{figure}
    \centering
    \includegraphics[width=0.7\linewidth]{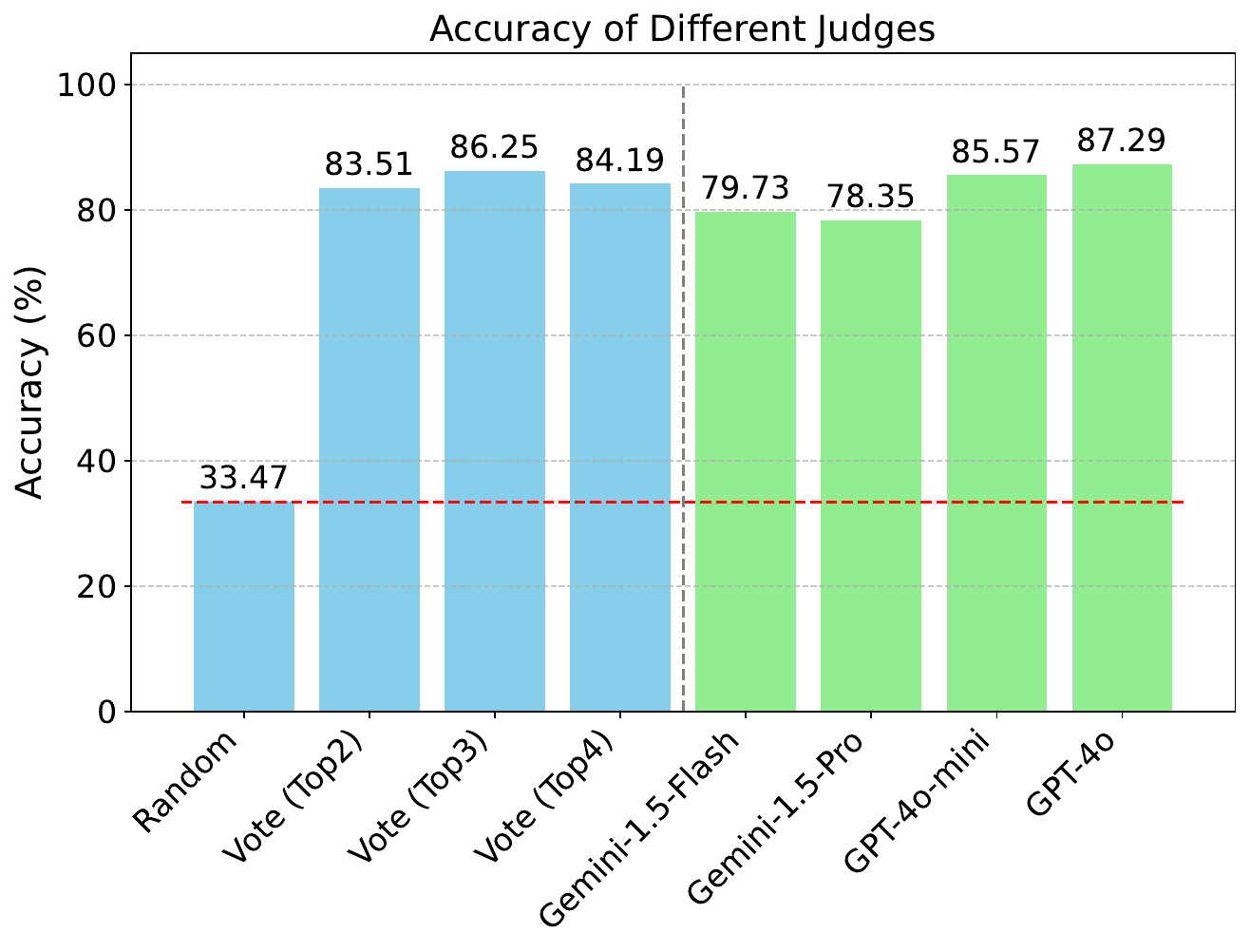}
    \vspace{-0.4cm}
    \caption{Evaluate the accuracy of various judging methods using human annotations as the gold standard. In the Vote (Top N) method, the top N models are used to cast votes.}
    \vspace{-0.4cm}
    \label{fig:judge_acc}
\end{figure}

\textbf{Automatic Judging.} To establish a ``gold standard'' for evaluating the accuracy of automatic judging, framed as a 3-choice task (A, B, or Tie), we created a benchmark guided by our judging criteria. Since no public human annotations are available for this, we engaged annotators to carefully evaluate a subset of battles. Annotation guidelines are detailed in the Appendix~\ref{app:annotation_judge}. Given the labor-intensive nature of this work—yielding about 7 annotations per hour—we randomly selected around 300 battles across various video lengths and models for annotation. Figure~\ref{fig:judge_acc} presents the accuracy comparison among SOTA LMMs and voting methods for automated judging. GPT-4o demonstrated the highest alignment with human preferences, achieving an 87.29\% agreement. Notably, employing a voting approach with multiple LMMs (Top2 to Top4) did not result in better agreement. As a result, we selected GPT-4o as the primary judge for determining the winning responses due to its strong alignment with human judgments.

\begin{figure}
    \centering
    \includegraphics[width=0.7\linewidth]{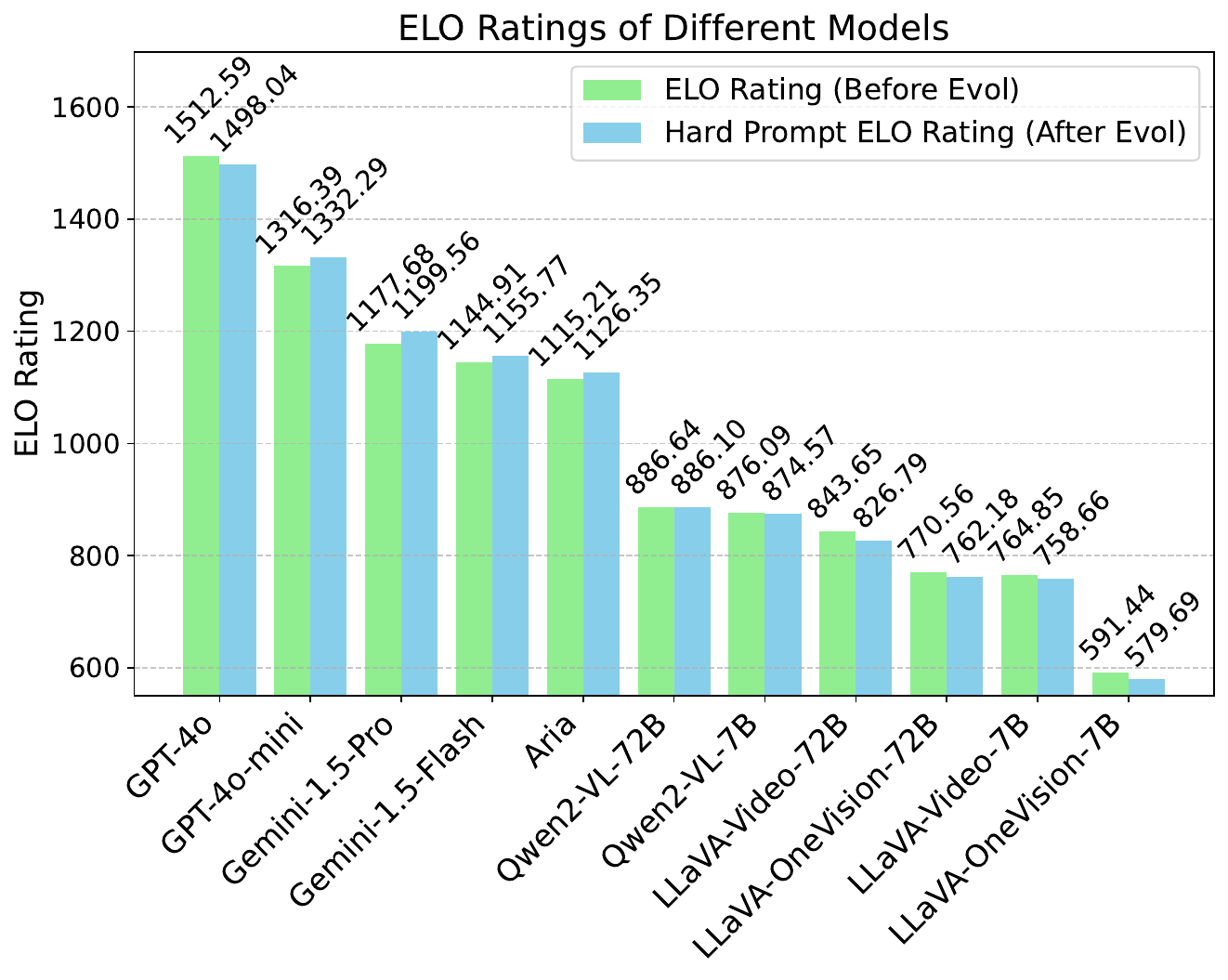}
    \vspace{-0.4cm}
    \caption{ELO ratings for models competing on questions before and after applying fault-aware evolution.}
    \vspace{-0.4cm}
    \label{fig:elo_diff}
\end{figure}
\begin{figure*}
    \centering
    \includegraphics[width=0.9\linewidth]{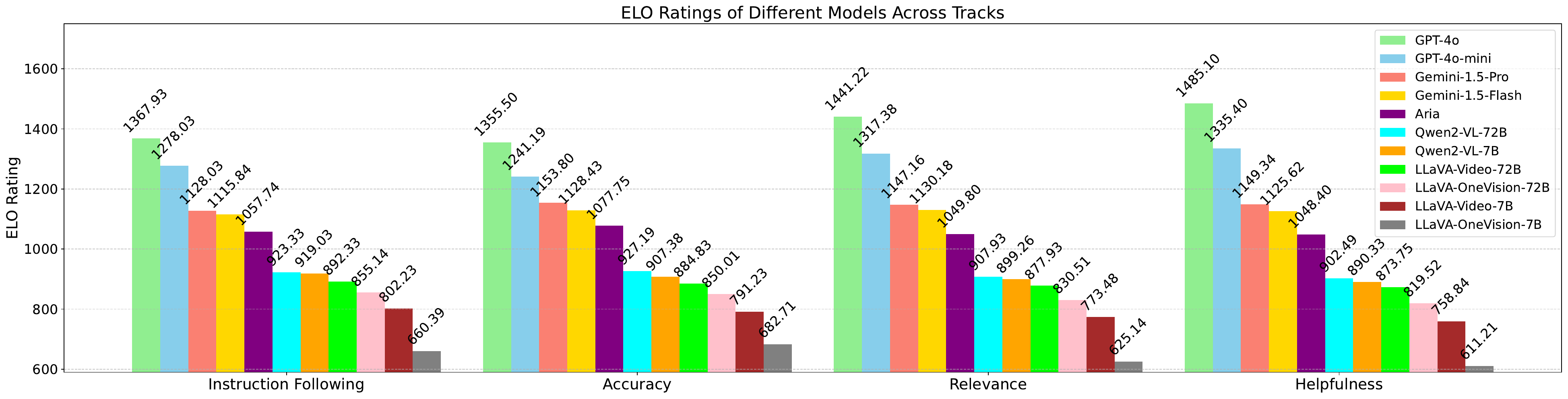}
    \vspace{-0.2cm}
    \caption{We evaluate the performance of various models based on four different judging standards.}
    \vspace{-0.2cm}
    \label{fig:elo_tracks}
\end{figure*}
\textbf{Leaderboard.} As shown in Table~\ref{tab:ELO}, we present the ELO ratings and win rates across 4 video length categories for 11 SOTA LMMs in video analysis. Our evaluation involves a total of 12,479 head-to-head battles, with each model participating in roughly 1,600 battles, and each model pair competing approximately 150 times—significantly surpassing the scale of WildVision Video Arena and demonstrating the scalability of our method. In terms of ELO ratings, the leading open-source LMM, Aria, lags behind the top proprietary LMM, GPT-4o, by a notable margin (-385.7), underscoring the benchmark's strong discriminative power. Unlike in VideoMME, where the score gap among the top six models is less than 10\%. For shorter videos (under 60 seconds), the gap between the top proprietary and open-source LMMs narrows, with Aria even surpassing the well-known Gemini-1.5-Pro on videos between 15 and 60 seconds. However, as video length increases, performance gaps widen considerably; for instance, Qwen2-VL sees a drop of over 100 ELO points, while Gemini-1.5-Flash gains around 180 ELO points when comparing results on short versus long videos. Notably, Aria demonstrates stable and strong performance across varying video lengths.

In addition, Figure~\ref{fig:elo_diff} presents the ELO ratings for models competing on questions before and after applying fault-aware evolution. When challenged with more difficult prompts, all open-source LMMs except Aria exhibit lower scores, whereas all proprietary LMMs, apart from GPT-4o, show improved scores. Figure~\ref{fig:elo_tracks} further details the performance of various LMMs across different judging standards. Notably, the gap between proprietary and open-source LMMs is most evident in the helpfulness and relevance tracks, compared to the instruction-following and accuracy tracks. This gap suggests that many open-source LMMs are primarily optimized for traditional, ability-focused benchmarks like VideoMME, which overlook user-centered aspects such as contextual relevance and information helpfulness. By focusing on a more comprehensive evaluation, our VideoAutoArena bridges this gap, providing deeper insights into the limitations and future development potential of LMMs. In the Appendix~\ref{app:example_response_judge}, we also include examples of different LMMs' responses as case studies to better understand the weaknesses of these models.

\begin{figure}
    \centering
    \vspace{-0.4cm}
    \includegraphics[width=\linewidth]{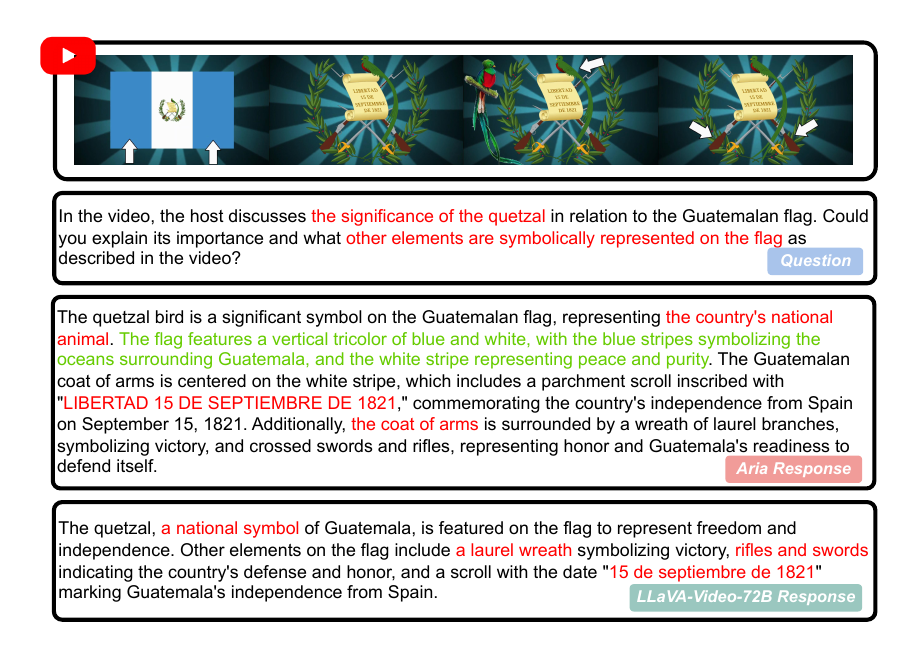}
    \vspace{-0.7cm}
    \caption{Example of a battle between Aria and LLaVa-Video-72B. \textcolor{red}{Red} highlights key content, while \textcolor{deepgreen}{green} highlights important details mentioned only by Aria.}
    \vspace{-0.5cm}
    \label{fig:case}
\end{figure}
\textbf{Battle Example.} As shown in Figure~\ref{fig:case}, we present an example of a battle between Aria and LLaVa-Video-72B. Both models follow the question and provide accurate information about the Guatemalan flag. However, only Aria includes additional details, such as the flag's vertical tricolor, while LLaVa-Video's response lacks sufficient depth on these aspects. Consequently, Aria's response is more helpful for users seeking to understand the flag's symbolism. The automated judge also selects Aria as the winner. Traditional multiple-choice QA benchmarks, like VideoMME, cannot reveal such open-ended limitations in LMMs. More examples are included in the Appendix~\ref{app:example_response_judge}.
\section{VideoAutoBench}\label{sec:bench}

While VideoAutoArena offers a novel approach to evaluating LMMs in video analysis, it is less immediately user-friendly than multiple-choice benchmarks like VideoMME and LongVideoBench, which provide straightforward scores based on model responses. In contrast, VideoAutoArena requires the target model to engage in comparative battles with other models to generate results. To streamline this evaluation process, we introduce \textbf{VideoAutoBench}, which combines the user-centric assessment strengths of VideoAutoArena with the simplicity and speed of traditional benchmarks. In our automated judging experiments, we included human annotators to label winners for a subset of battles, using these questions and non-tied responses as reference answers. In VideoAutoBench, we employ GPT-4o as the judging model to evaluate each model's responses against the human-selected or rejected answers, with GPT-4o voting based on the same standards used in VideoAutoArena. When competing with human-selected answers, a model earns 1 point for a win, 0.5 for a tie, and 0 for a loss, with the final score being the average across all battles. When competing with human-rejected answers, only a win earns 1 point. Table~\ref{tab:videoautobench} presents the performance of different LMMs, showing the similar rank as in VideoAutoArena.

\begin{table}
    \small
    \centering
    \begin{tabular}{lccc}
        \toprule
        \textbf{Models} & \textbf{vs. Sel.}&  \textbf{vs. Rej.} & \textbf{Avg.}\\
        \midrule
        \Openaiemoji{}~\textbf{GPT-4o} & \textbf{70.98} & \textbf{94.12} & \textbf{82.55}\\
        \Openaiemoji{}~\textbf{GPT-4o-mini} & 49.80 & 92.16 & 70.98\\
        \Googleemoji{}~\textbf{Gemini-1.5-Pro} & 28.24 & 82.74 & 55.49\\
        \Googleemoji{}~\textbf{Gemini-1.5-Flash} & 27.25 & 81.96 & 54.61\\
        \Ariaemoji{}~\textbf{Aria} & \textbf{19.80} & \textbf{76.86} & \textbf{48.33}\\
        \Qwenemoji{}~\textbf{Qwen2-VL-72B} & 13.92 & 64.71 & 39.32\\
        \Qwenemoji{}~\textbf{Qwen2-VL-7B} & 11.96 & 60.00 & 35.98\\
        \llavanextmoji{}~\textbf{LLaVA-Video-72B} & 7.45 & 56.08 & 31.77\\
        \llavanextmoji{}~\textbf{LLaVA-OneVision-72B} & 4.12 & 52.16 & 28.14\\
        \llavanextmoji{}~\textbf{LLaVA-Video-7B} & 5.29 & 46.67 & 25.98\\
        \llavanextmoji{}~\textbf{LLaVA-OneVision-7B} & 3.53 & 30.98 & 17.26\\
        \bottomrule
    \end{tabular}
    \caption{LMMs compete against human selected or rejected answers in our \textbf{VideoAutoBench}.}
    \vspace{-0.4cm}
    \label{tab:videoautobench}
\end{table}

\section{Conclusion}

We introduce VideoAutoArena, an automated arena-style benchmark that addresses the limitations of traditional multiple-choice video QA benchmarks. Using user simulation, peer battles, automated judging, and fault-driven evolution, VideoAutoArena enables a scalable, user-centric evaluation for complex video analysis tasks. Alongside, we present VideoAutoBench, a streamlined evaluation comparing model responses to human-labeled answers. Experiments on 11 SOTA LMMs reveal a notable performance gap between closed and open-source models, particularly for long videos, challenging questions, and scenarios involving user background relevance and response helpfulness.
{
    \small
    \bibliographystyle{ieeenat_fullname}
    \bibliography{main}
}

\clearpage
\appendix
\section{Prompts}
\begin{figure*}
    \centering
    \includegraphics[width=0.92\textwidth]{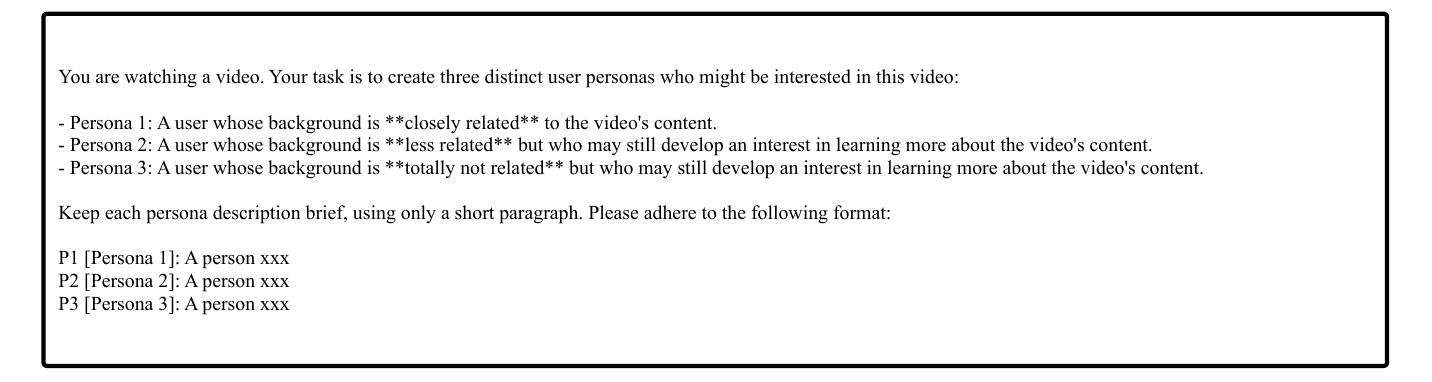}
    \caption{The prompt for video content-constrained persona generation.}
    \label{fig:persona_prompt}
\end{figure*}
\begin{figure*}
    \centering
    \includegraphics[width=0.92\textwidth]{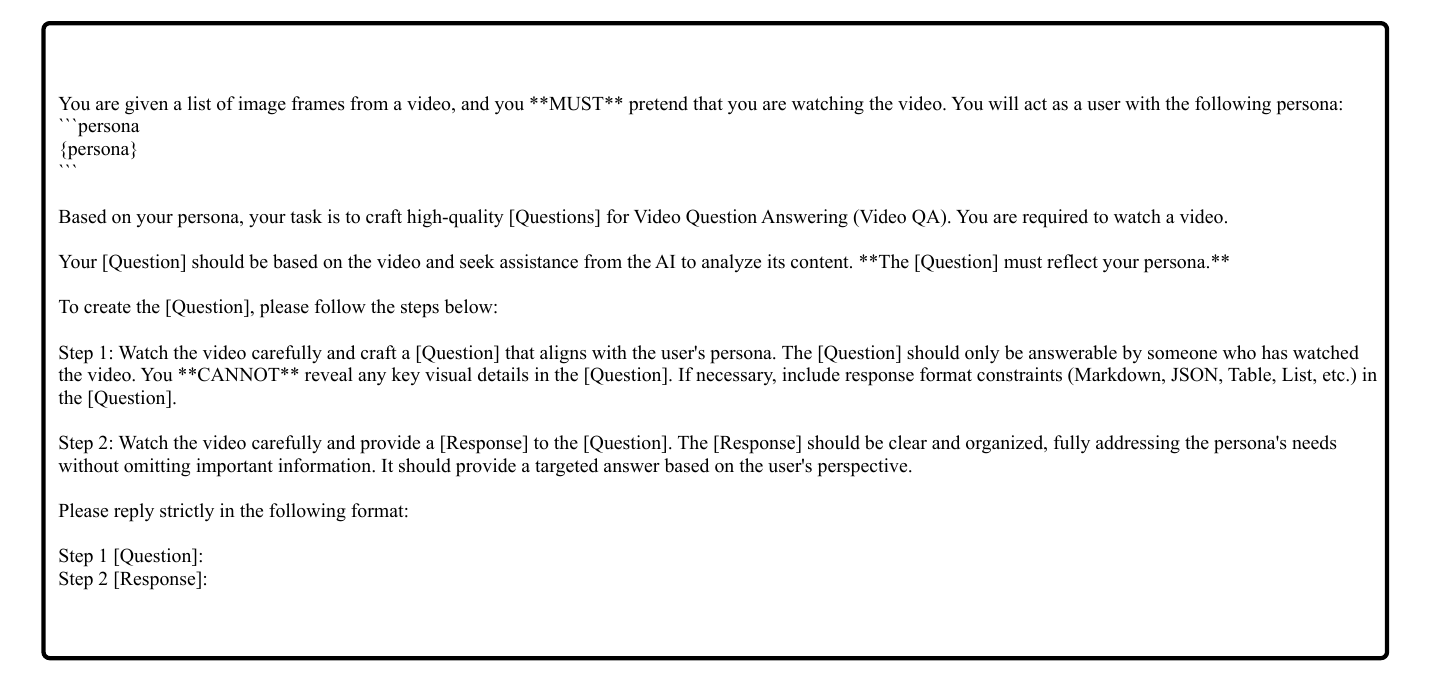}
    \caption{The prompt for persona-constrained video question asking.}
    \label{fig:question_prompt}
\end{figure*}
\subsection{User Simulation}\label{app:user}

In Figure~\ref{fig:persona_prompt} and~\ref{fig:question_prompt}, we include the prompts for video content-constrained persona generation and persona-constrained video question asking.

\begin{figure*}
    \centering
    \includegraphics[width=0.95\textwidth]{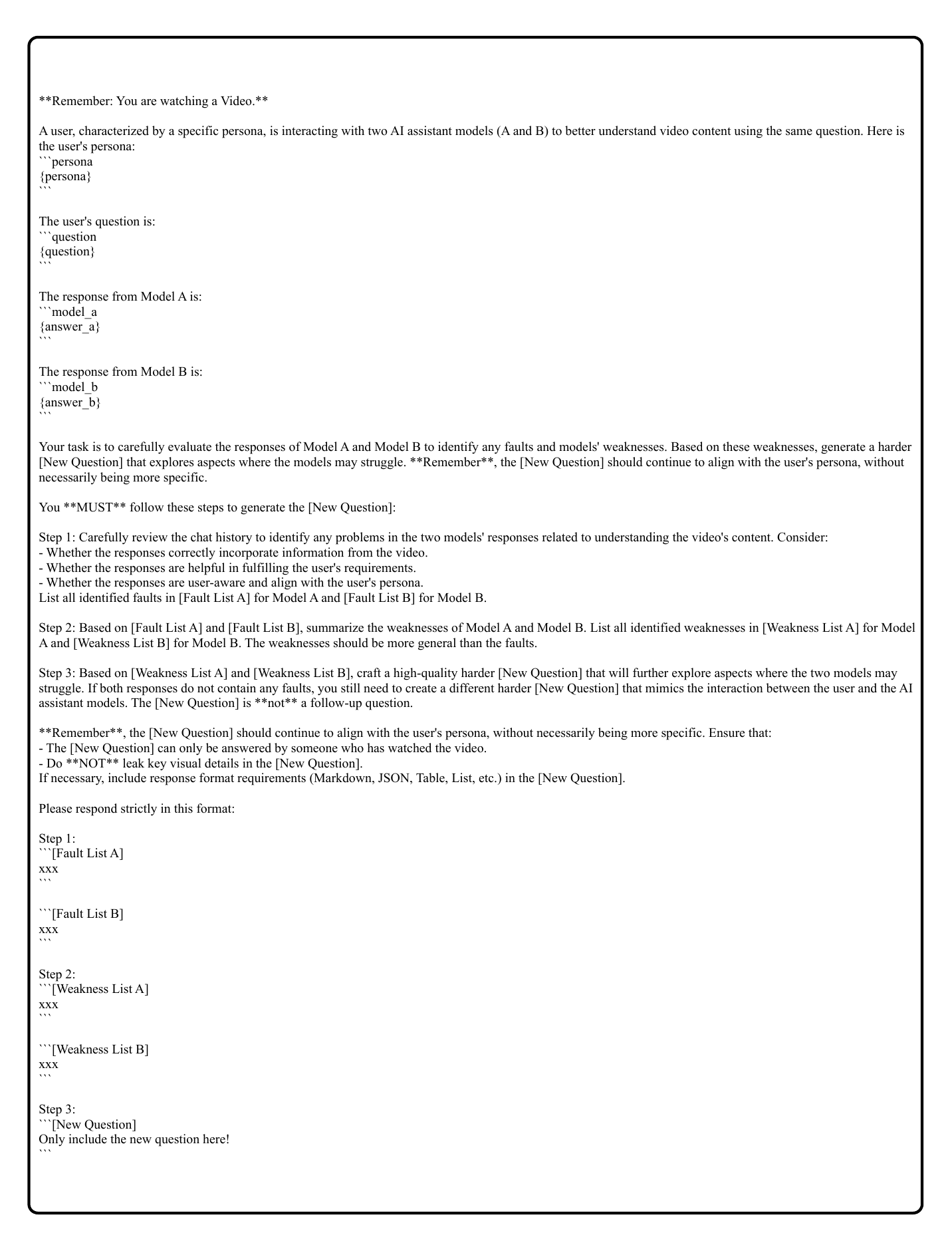}
    \caption{The prompt for our fault-driven evolution generates new questions based on the responses from the two models.}
    \label{fig:evol_prompt}
\end{figure*}
\subsection{Fault-Driven Evolution}\label{app:evol}

Figure~\ref{fig:evol_prompt} includes the prompt for the fault-driven evolution.

\subsection{Automatic Judging}\label{app:judge}

\begin{figure*}
    \centering
    \includegraphics[width=0.95\textwidth]{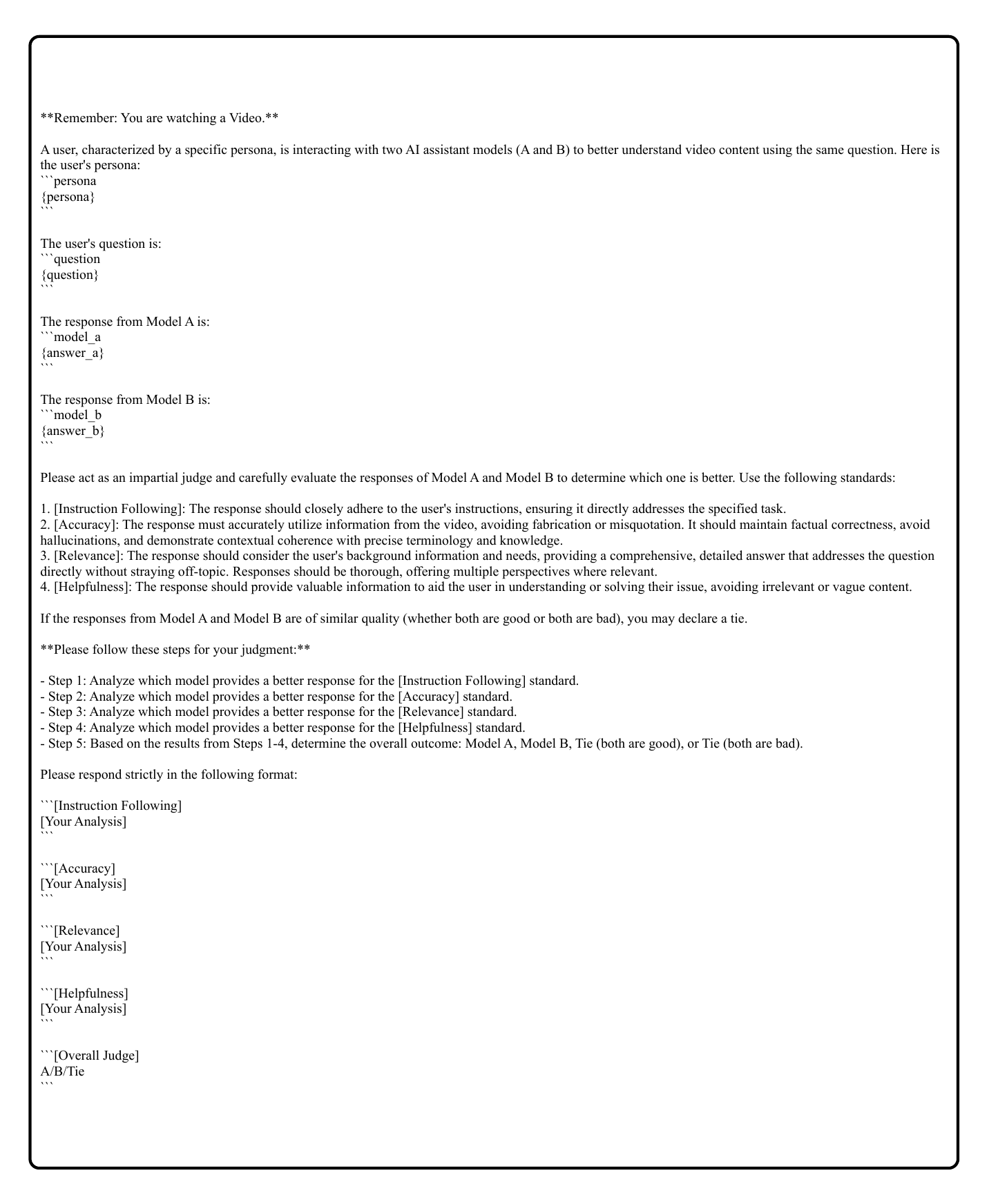}
    \caption{The prompt for our automatic judging.}
    \label{fig:judge_prompt}
\end{figure*}

Figure~\ref{fig:judge_prompt} includes the prompt for the automatic judging. Our VideoAutoBench adopts the same prompt for judging.

\begin{figure*}
    \centering
    \includegraphics[width=0.95\textwidth]{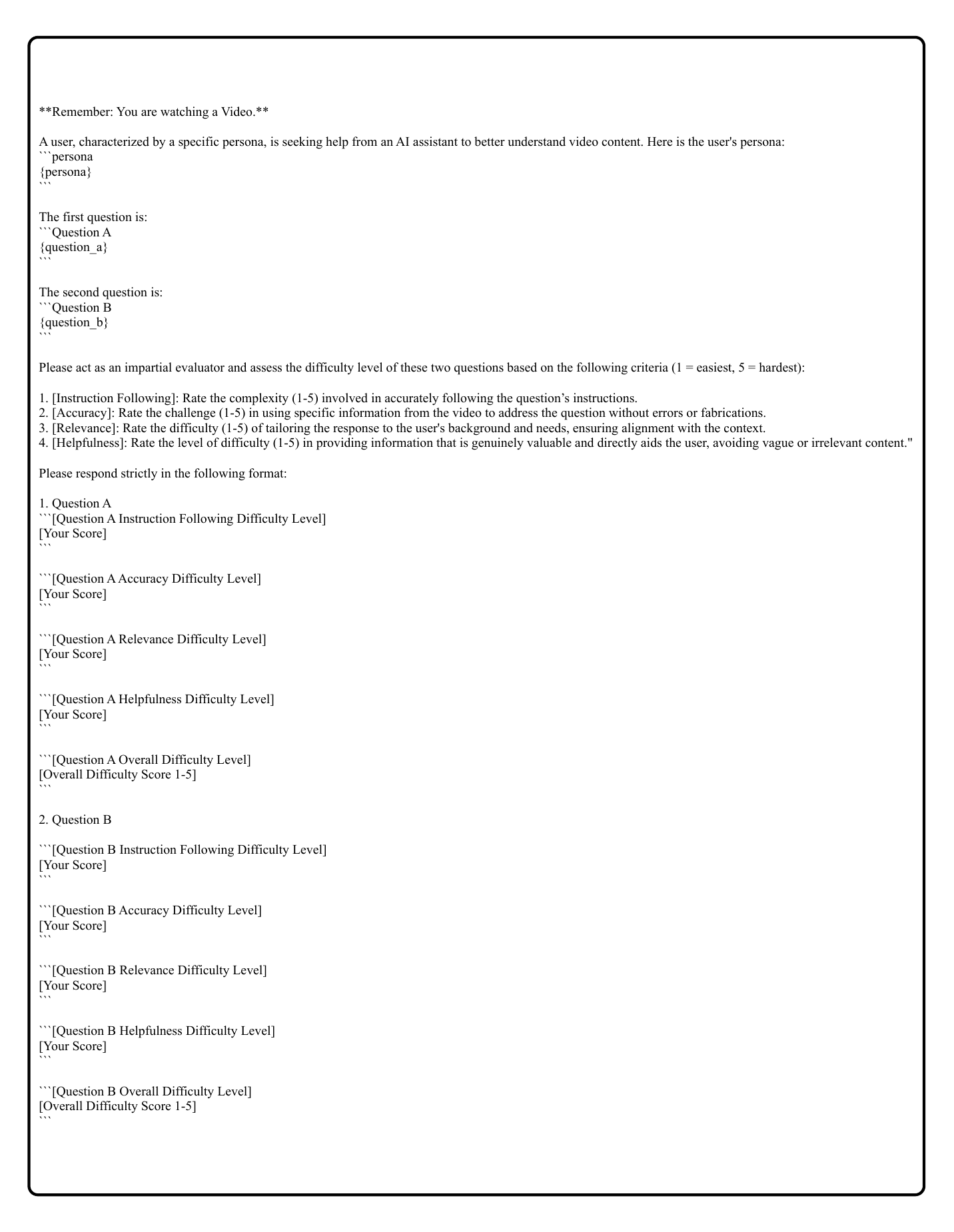}
    \caption{The prompt for question complexity evaluation.}
    \label{fig:complex_prompt}
\end{figure*}
\subsection{Difficulty Level Evaluation}\label{app:comp}

Figure~\ref{fig:complex_prompt} includes the prompt for the complexity evaluation.

\section{Experimental Details}\label{app:exp}

\subsection{Statistics}
In VideoAutoArena, Model A wins 5,620 battles, Model B wins 5,941 battles, and 918 ties. VideoAutoBench consists of 255 samples corresponding to 244 unique videos, with an average duration of 478.5 seconds. The duration distribution includes 62 videos for 8-15s, 62 for 15-60s, 60 for 180-600s, and 60 for 900-3,600s. The samples span 10 categories: 29 from Movies, 50 from Life Vlogs, 9 from Geography, 13 from History, 12 from News Programs, 9 from Art, 6 from STEM, 8 from Computer Science, 55 from Cooking Recipes, and 53 from Travel Guides.

\subsection{LMMs Selection}

In our experiments, we evaluate 11 SOTA LMMs:
\begin{enumerate}
    \item \texttt{gpt-4o-2024-05-13},
    \item \texttt{gpt-4o-mini-2024-07-18},
    \item \texttt{gemini-1.5-pro},
    \item \texttt{gemini-1.5-flash},
    \item \texttt{rhymes-ai/Aria},
    \item \texttt{Qwen/Qwen2-VL-72B-Instruct},
    \item \texttt{Qwen/Qwen2-VL-7B-Instruct}
    \item \texttt{lmms-lab/LLaVA-Video-72B-Qwen2},
    \item \texttt{lmms-lab/LLaVA-Video-7B-Qwen2},
    \item \texttt{lmms-lab/llava-onevision-qwen2-72b-ov},
    \item \texttt{lmms-lab/llava-onevision-qwen2-7b-ov}.
\end{enumerate}
We selected the top 10 LMMs that support long video analysis, along with their smaller versions, based on their performance on VideoMME as of October 15, 2024. Additional open-source LMMs were excluded for two reasons: first, some were released concurrently with our work, leaving insufficient time for evaluation; second, others exhibited weak performance on VideoMME or lacked support for long video analysis. Consequently, our experiments are limited to the 11 most popular and SOTA LMMs.

For user simulation, fault-driven evolution, automatic judging, and difficulty level evaluation, we use \texttt{gpt-4o-2024-08-06}, ensuring that the examiner and judge remain distinct from the LMMs throughout the entire evaluation process.

\subsection{Hyperparameters}

For user simulation, fault-driven evolution, automatic judging, and difficulty level evaluation, each video is uniformly sampled into a maximum of 128 frames, while response generation uses up to 64 frames. Each frame is resized to $512 \times 512$. For API-based LMMs, the \texttt{max\_tokens} parameter is set to 4096, with other settings using default values. For open-source LMMs, the temperature is set to 0.7, and \texttt{max\_new\_tokens} is limited to 2048.

\section{Examples}\label{app:example}
\begin{figure*}
    \centering
    \includegraphics[width=\textwidth]{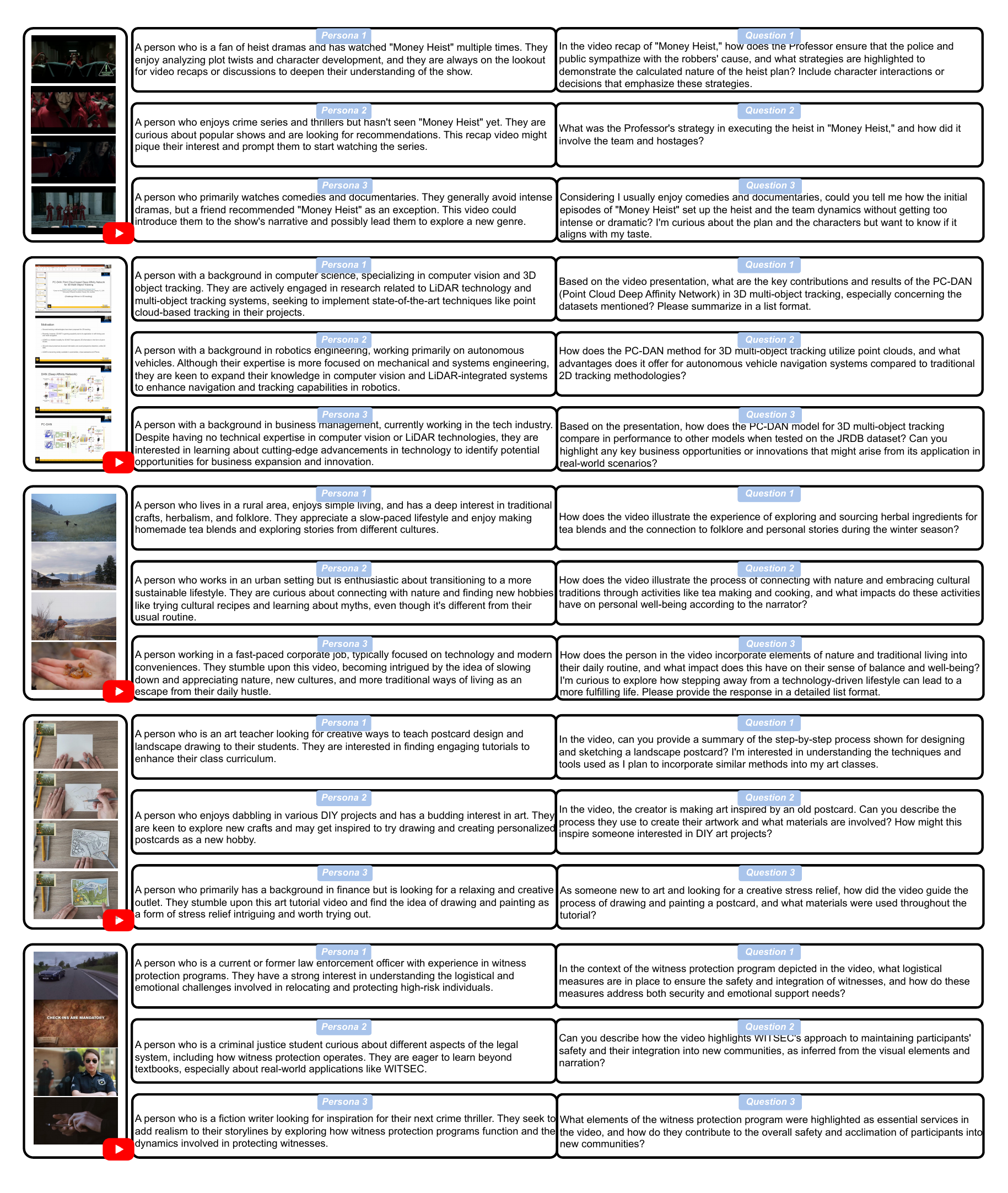}
    \caption{Examples of our user simulation include five videos from diverse domains: Movies, Computer Science, Life Vlogs, Art, and News Programs. To save space, we only showcase 4 frames of each video.}
    \label{fig:example_5_user_sim}
\end{figure*}
\subsection{Persona}\label{app:example_persona}

As shown in Figure~\ref{fig:example_5_user_sim}, we include 15 examples of different personas for 5 different videos.

\subsection{Question}\label{app:example_question}

As shown in Figure~\ref{fig:example_5_user_sim}, we include 15 examples of different questions for 5 different videos.

\begin{figure*}
    \centering
    \includegraphics[width=\textwidth]{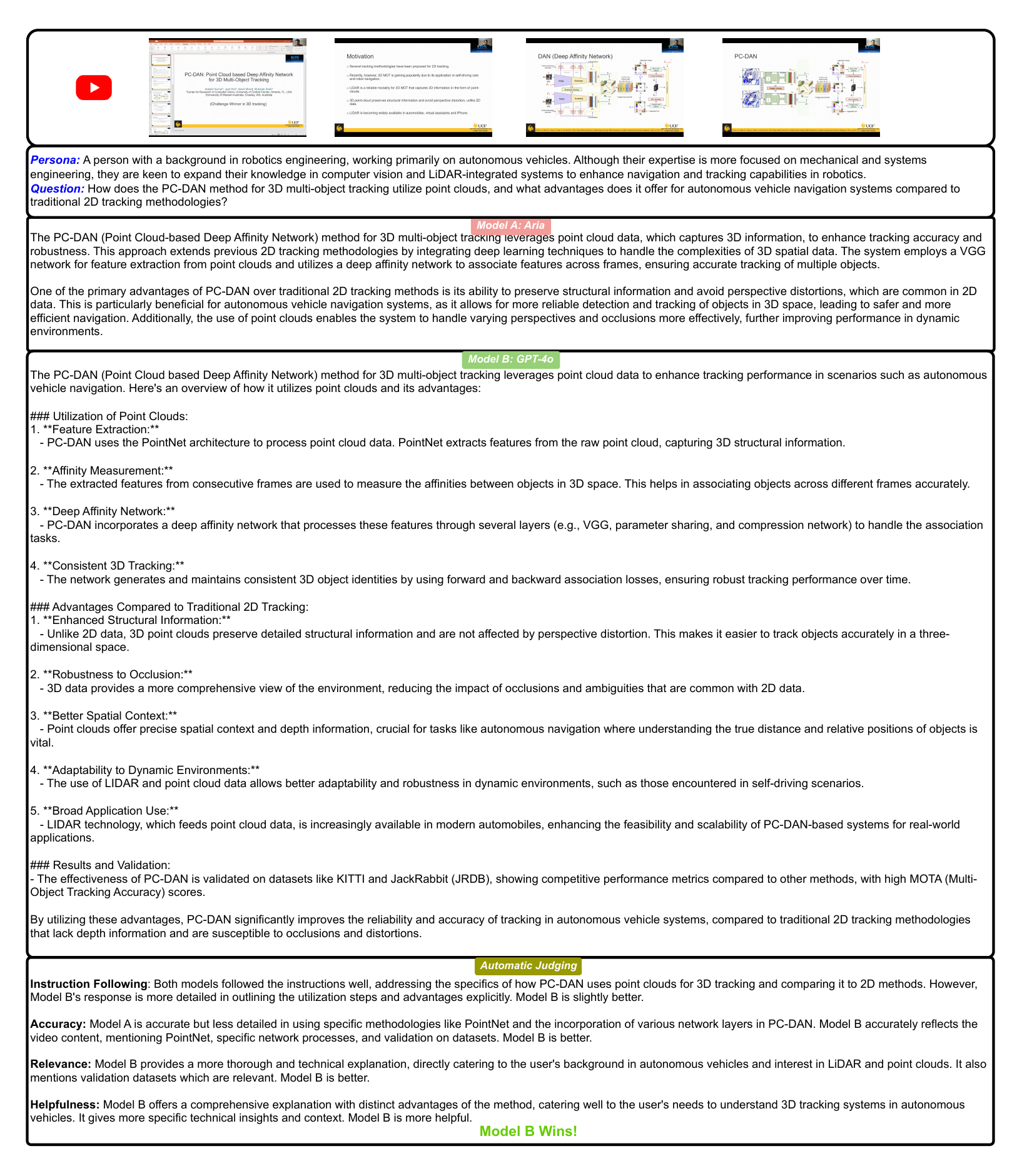}
    \caption{Example of the battle between Aria and GPT-4o.}
    \label{fig:res_judge_1}
\end{figure*}

\begin{figure*}
    \centering
    \includegraphics[width=\textwidth]{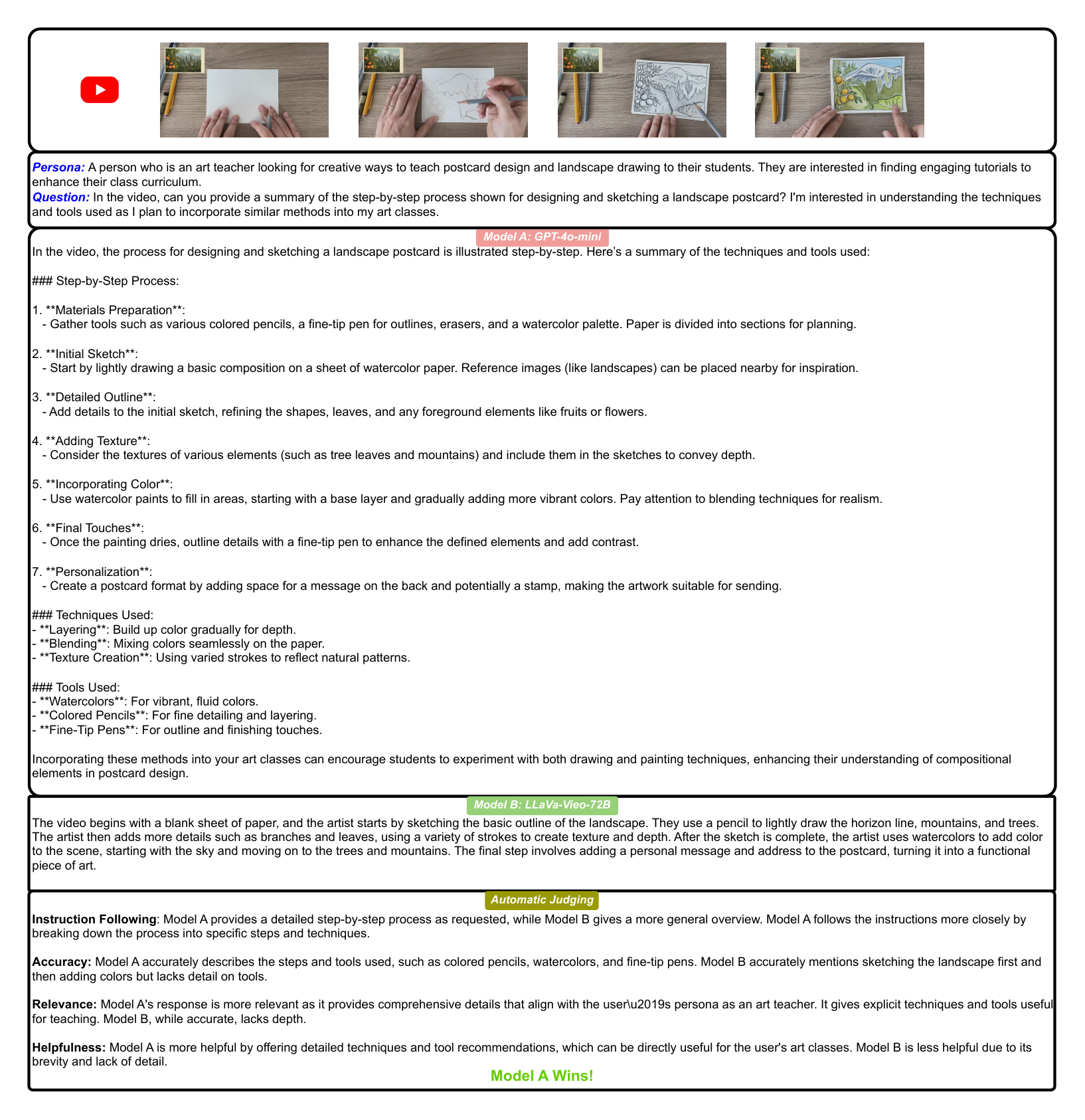}
    \caption{Example of the battle between GPT-4o-mini and LLaVa-Video-72B.}
    \label{fig:res_judge_2}
\end{figure*}

\begin{figure*}
    \centering
    \includegraphics[width=\textwidth]{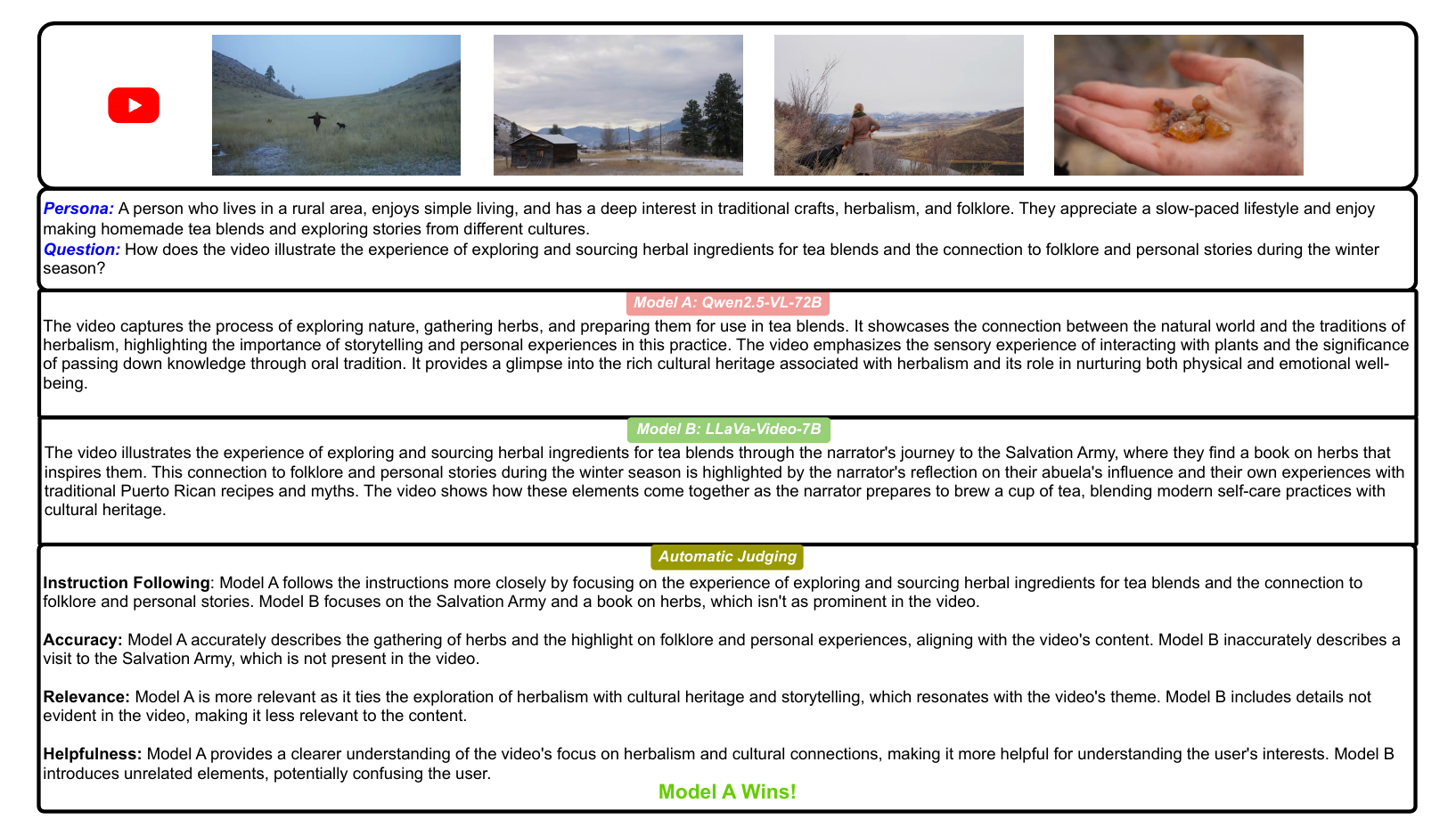}
    \caption{Example of the battle between Qwen2-VL-72B and LLaVa-Video-7B.}
    \label{fig:res_judge_3}
\end{figure*}

\begin{figure*}
    \centering
    \includegraphics[width=\textwidth]{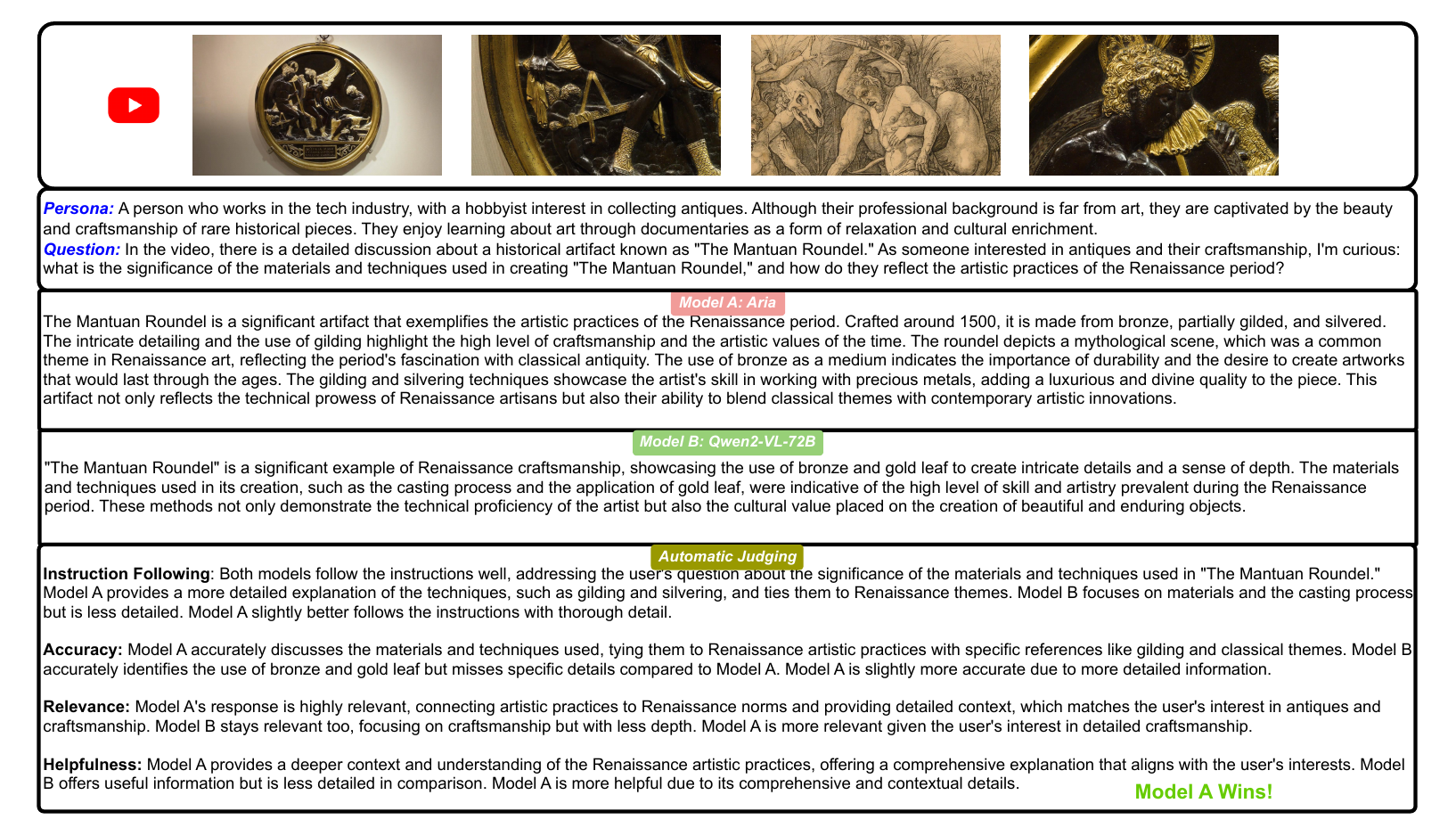}
    \caption{Example of the battle between Aria and Qwen2-VL-72B.}
    \label{fig:res_judge_4}
\end{figure*}
\subsection{Responses and Judging}\label{app:example_response_judge}

As shown in Figure~\ref{fig:res_judge_1},~\ref{fig:res_judge_2},~\ref{fig:res_judge_3}, and~\ref{fig:res_judge_4}, we include the battle examples between different models.

\section{Human Annotations}\label{app:annotation}

\subsection{Question Ranking}\label{app:annotation_rank}

\begin{figure*}
    \centering
    \includegraphics[width=\textwidth]{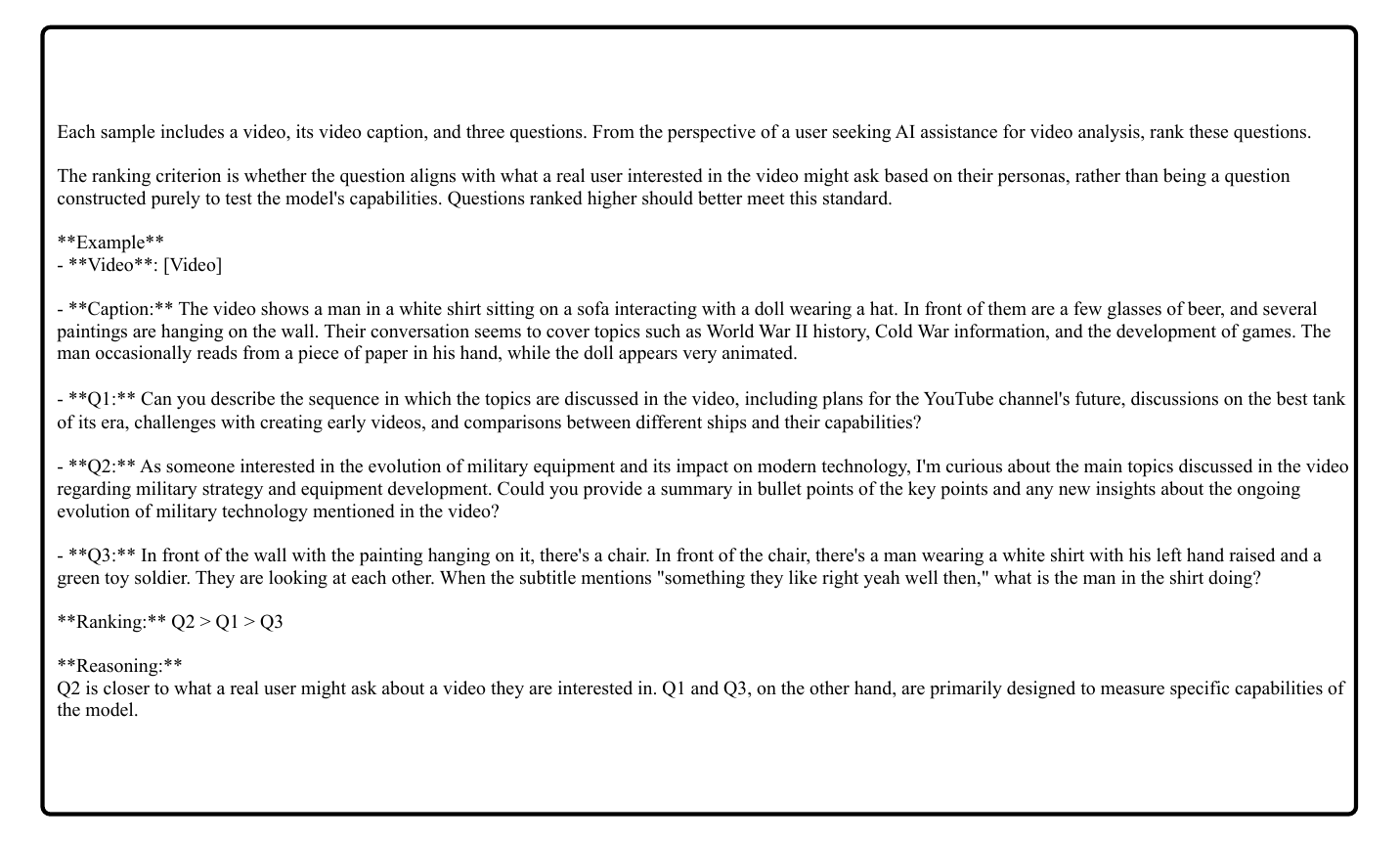}
    \caption{The guideline for the question ranking annotation.}
    \label{fig:ranking}
\end{figure*}

\begin{figure*}
    \centering
    \includegraphics[width=\textwidth]{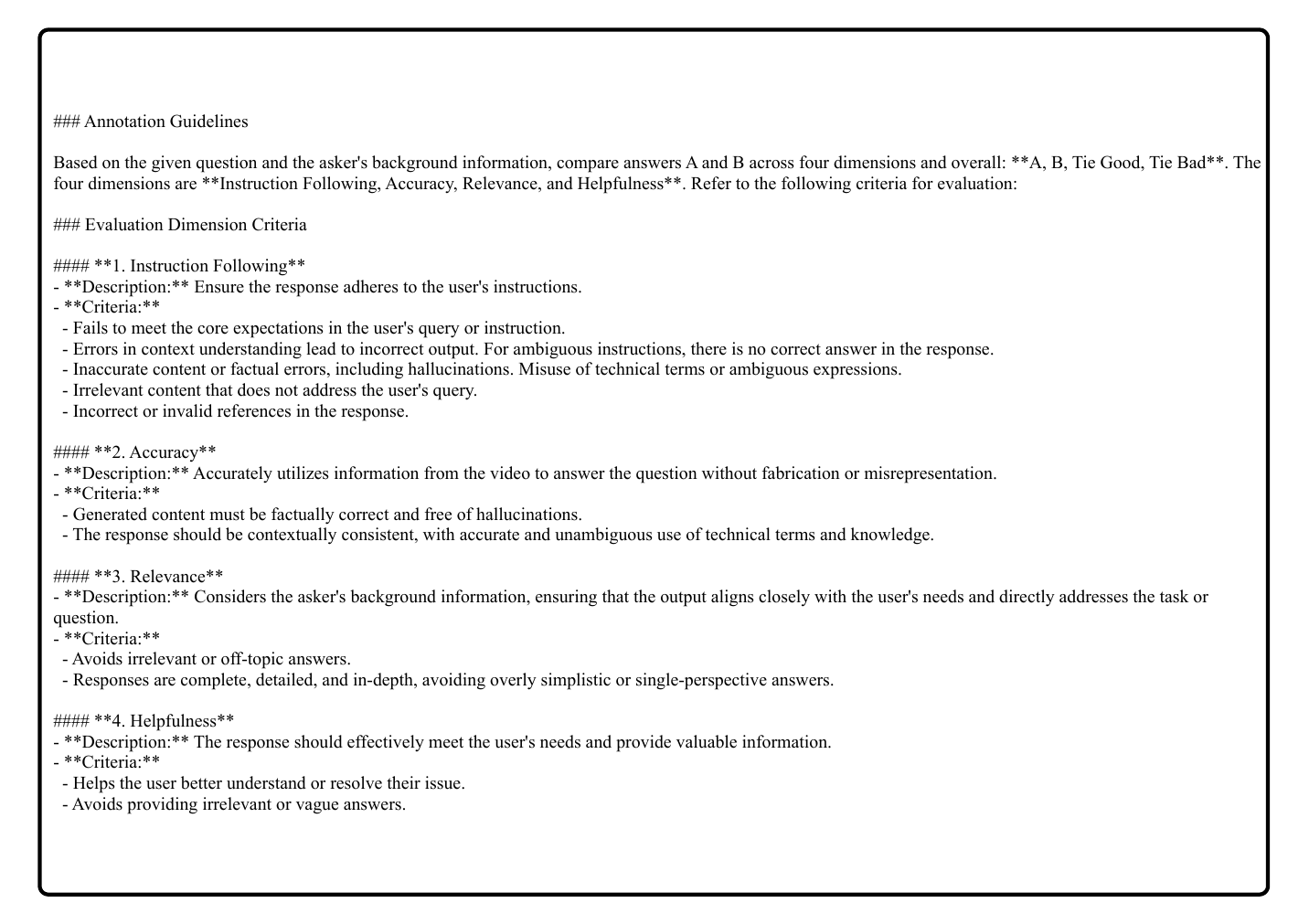}
    \caption{The guideline for the response judging annotation.}
    \label{fig:judging}
\end{figure*}
In Figure~\ref{fig:ranking}, we include our guideline for the question ranking annotation.

\subsection{Judging}\label{app:annotation_judge}

In Figure~\ref{fig:judging}, we include our guideline for the response judging annotation.

\section{Limitation}

VideoAutoArena and VideoAutoBench currently lack evaluations for multi-turn and non-English interactions, primarily due to the limited multi-turn conversational capabilities and restricted non-English proficiency of current open-sourced LMMs. Moreover, the automatic judging system tends to favor detailed responses, a preference also observed in human evaluations. While detailed responses are often more helpful to users, this introduces challenges in evaluating LMMs. We implemented the style-control method from LMSYS Chatbot Arena to adjust ELO ratings by penalizing stylistic factors. However, we found this approach unsuitable for evaluating current LMMs. For example, while Aria and Qwen2-VL-72B outperform Gemini-1.5-Pro in ELO ratings, Gemini-1.5-Pro consistently achieves significantly higher win rates. Manual review of Gemini-1.5-Pro's outputs revealed that its responses are not only more detailed but also of higher quality compared to those from the two open-source LMMs. Additionally, most open-source LMMs tend to generate less detailed responses than proprietary LMMs, causing the style-control mechanism to disproportionately penalize proprietary models. This imbalance leads to unfair evaluations. To address this issue, a more effective style-control mechanism should ensure that competing LMMs produce responses with comparable levels of detail, thereby enabling a fairer evaluation.

To address these limitations, future work will focus on expanding VideoAutoArena and VideoAutoBench to include multiturn and multilingual data. Additionally, we aim to refine our definitions of user simulation, developing a systematic approach for generating battles that encompass a broader and more inclusive range of scenarios while maintaining high separability and alignment with human judgment. Furthermore, we plan to explore advanced style-control techniques and unbiased LMMs-as-judge to further enhance the robustness and fairness of our LMM-based evaluation framework.

\end{document}